\pdfoutput=1

\documentclass[11pt]{article}

\usepackage{acl}

\usepackage{times}
\usepackage{latexsym}
\usepackage{natbib}
\usepackage[normalem]{ulem}

\usepackage[T1]{fontenc}

\usepackage[utf8]{inputenc}

\usepackage{microtype}

\usepackage{inconsolata}

\usepackage{booktabs}       
\usepackage{amsfonts}       
\usepackage{nicefrac}       
\usepackage{xcolor}         
\usepackage{enumitem}
\usepackage{multicol}
\usepackage{multirow}
\usepackage{array}
\usepackage{graphicx}
\usepackage{wrapfig}
\usepackage{latexsym}
\usepackage{amsmath}
\usepackage{listings}
\usepackage{mdframed}
\mdfsetup{nobreak=true}
\usepackage{titletoc}

\newcommand\benchmarkName{\textbf{IL-TUR}}
\newcommand\website{\url{https://exploration-lab.github.io/IL-TUR/}}

\newenvironment{alltt}{\ttfamily}{\par}

\title{\benchmarkName: Benchmark for Indian Legal Text Understanding and Reasoning}

\author{{\bf Abhinav Joshi}$^\mathparagraph$\thanks{\ \ Equal Contributions} \qquad
{\bf Shounak Paul}$^\diamond$\footnotemark[1] \\ 
{\bf Akshat Sharma}$^\mathparagraph$ \qquad 
\textbf{Pawan Goyal}$^\diamond$ \qquad \textbf{Saptarshi Ghosh}$^\diamond$ \\ {\bf Ashutosh Modi}$^\mathparagraph$\thanks{\ \ Corresponding Author}  
 \\ 
        $^\mathparagraph$IIT Kanpur, 
        $^\diamond$IIT Kharagpur\\
  \texttt{shounakpaul95@kgpian.iitkgp.ac.in}, \texttt{\{pawang,saptarshi\}@cse.iitkgp.ac.in}, \\
  \texttt{\{ajoshi, akshatsh, ashutoshm\}@cse.iitk.ac.in}  
}

\begin{document}
\maketitle

\begin{abstract}
Legal systems worldwide are inundated with exponential growth in cases and documents. There is an imminent need to develop NLP and ML techniques for automatically processing and understanding legal documents to streamline the legal system. However, evaluating and comparing various NLP models designed specifically for the legal domain is challenging. This paper addresses this challenge by proposing \benchmarkName: Benchmark for Indian Legal Text Understanding and Reasoning. \benchmarkName\  contains monolingual (English, Hindi) and multi-lingual (9 Indian languages) domain-specific tasks that address different aspects of the legal system from the point of view of understanding and reasoning over Indian legal documents. We present baseline models (including LLM-based) for each task, outlining the gap between models and the ground truth. To foster further research in the legal domain, we create a leaderboard (available at: \url{https://exploration-lab.github.io/IL-TUR/}) where the research community can upload and compare legal text understanding systems. 
\end{abstract}

\noindent\textit{``Justice delayed is justice denied''} - Legal Maxim

\vspace{-1mm}
\section{Introduction} \label{sec:intro}
\vspace{-1mm}


\noindent Besides several other purposes, legal systems have been established in various countries to ensure, at the very minimum, order and fairness in society and to safeguard fundamental human rights. However, legal systems worldwide struggle with exponentially growing legal cases in various courts. It is even more pronounced in populous countries; e.g., in India, there are about $50$ million pending cases in multiple courts at various levels (district, state, federal) \citep{njdc-district}. Such a massive backlog of cases goes against the fundamental human right of fair access to justice. Documents in different natural languages are the backbone of various legal processes. Natural Language Processing (NLP) based techniques could be helpful in various legal processes involving fundamental tasks related to information extraction, document understanding, and prediction. This paper introduces \benchmarkName,  a benchmark for \textit{Indian Legal Text Understanding and Reasoning}. The purpose of \benchmarkName\  is twofold. First, it aims to foster research in the Legal-NLP (L-NLP) domain and plans to address the pain points associated with processing legal texts (see below); second, it provides a platform for comparing different models and further advancing the L-NLP domain.    

\begin{figure}[t]
  \centering
  \vspace{-3mm}
 \includegraphics[scale=0.26]{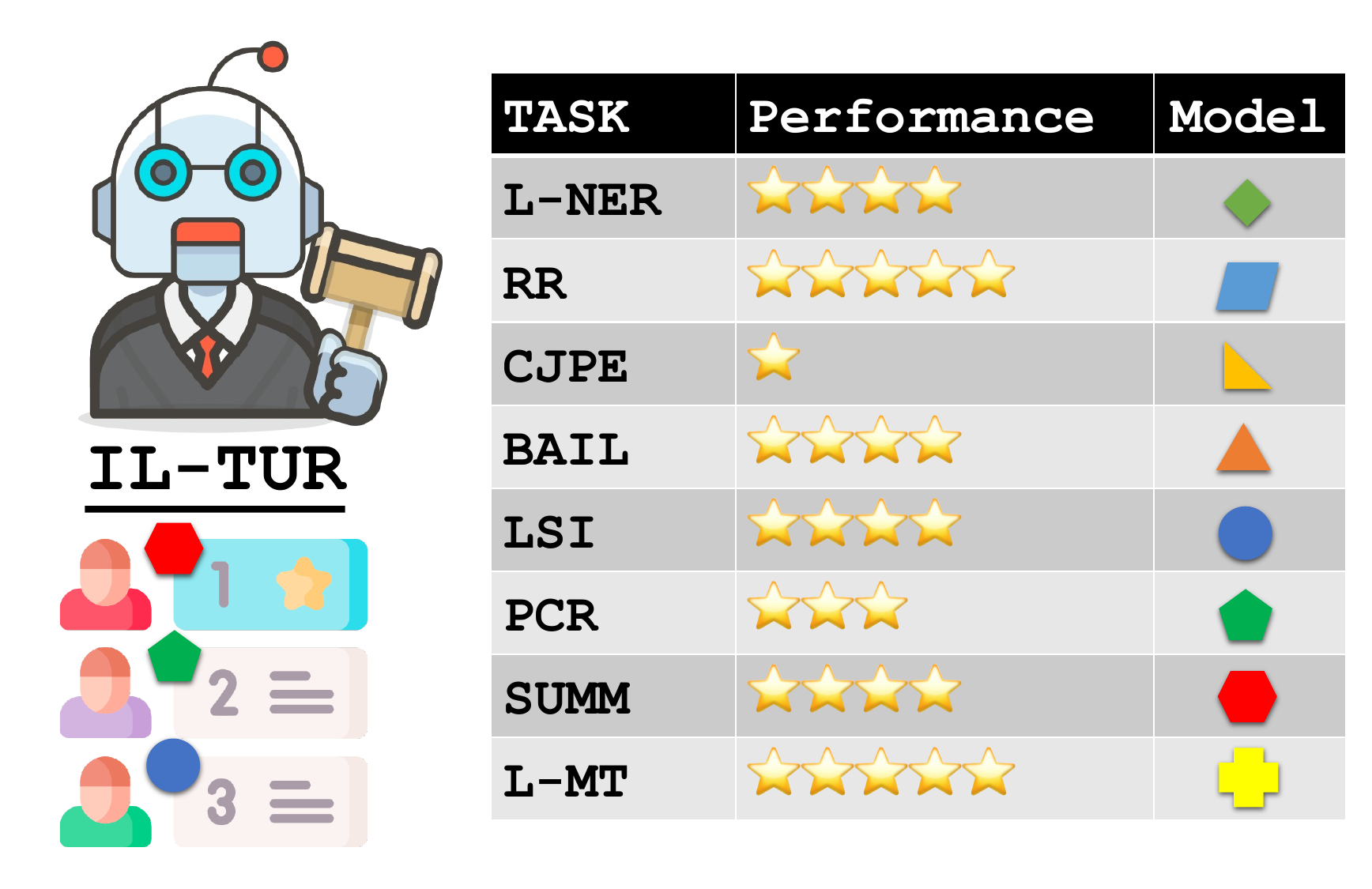}
  \caption{\benchmarkName: A consolidated benchmark covering a wide range of legal text understanding and reasoning tasks with a publically available leaderboard.}
  \label{fig:il-tur}
  \vspace{-6mm}
\end{figure}

\noindent\textbf{Why a separate benchmark for the legal domain?} The legal text involves natural language but differs from the regular text used to train NLP models. 1) Many of the terms used in legal documents are domain-specific. For example, some words used in everyday language have specialized meanings in legal parlance. The presence of a different lexicon posits a need for specialized NLP tools to handle legal texts. 2) Legal documents are typically very long compared to regular texts. For example, the average length of a legal document from the Supreme Court of India (SCI) is $4000$ words \citep{malik2021ildc}. It poses a challenge for existing NLP models (e.g., LLMs) as the information is spread throughout the document and must be linked together for reasoning.
Moreover, many of the existing language models (e.g., BERT \citep{devlin-etal-2019-bert}) have limitations on the length ($512$ tokens) of the input. It requires developing specialized models for processing and handling long legal documents. 3) Legal documents are highly unstructured and sometimes noisy (for example, in the Indian setting, most documents are typed manually in the courts and prone to grammatical mistakes and typos). The absence of structure in the documents makes extracting semantically relevant information from large chunks of text difficult. 4) The legal domain is further subdivided into specialized sub-domains; for example, criminal law differs from civil law, and both differ from banking and insurance law. Even though some fundamental legal principles are shared across various laws, models trained on a particular law (e.g., civil law) may not work on another (e.g., banking and insurance law). Hence, domain adaptation is a challenge. 5) Lastly, many existing state-of-the-art (SOTA) NLP models are black boxes; however, explainability is not a second-class citizen for the legal domain. For models to be widely usable by legal practitioners, these need to be explainable. Due to the above reasons, a separate set of models/systems is required to process and understand legal documents. Given the huge backlog of cases, NLP-based technologies could come to our rescue and help streamline the legal workflow. Even a small technical intervention can have a considerable impact. Hence, a benchmark is needed to promote the development of models in this area. In a nutshell, we make the following contributions:
\begin{itemize}[nosep,noitemsep]
\item We introduce \benchmarkName: a benchmark for Indian Legal Text Understanding and Reasoning. The benchmark has eight tasks (in English and 9 Indian languages) requiring different types of legal knowledge and skills to solve. Moreover, the list of tasks is not exhaustive, and we plan to keep adding more tasks to \benchmarkName. Currently, there are various L-NLP-specific tasks; however, these occur in isolation, making it difficult to keep track of progress made in the field. Similar to existing NLP benchmarks (e.g., GLUE \citep{wang2018glue}), we consolidate and harmonize some of the existing L-NLP tasks and create new tasks resulting in a unified benchmark. 
\item We report baseline model results on each of the tasks. We also experiment with various LLMs (\S\ref{sec:models}), and results show that LLMs are far from solving the tasks and hence point towards the need to develop better models. 
\item We release the dataset and baseline models associated with each task. Further, we create a leaderboard where anyone can upload their model and test against the baselines and other proposed systems (e.g., Fig. \ref{fig:il-tur}). The datasets, models, and the leaderboard are available via the following website: \website. 
\end{itemize}

\section{Related Work} \label{sec:related}

Over the past few years, L-NLP has been a fertile area for research. Researchers have explored different aspects of the legal domain via various tasks such as Prior Case Retrieval \citep{joshi-etal-2023-ucreat,jackson2003information}, Case Prediction \citep{malik2021ildc,chalkidis-etal-2019-neural,strickson2020legal,kapoor-etal-2022-hldc}, Summarization \citep{moens1999abstracting}, Semantic Segmentation of Legal Documents \citep{malik-rr-2021,kalamkar-etal-2022-corpus,bhattacharya2019identification}, and Information Extraction and Retrieval \citep{tran2019building,lagos2010event}. On the modeling side, various techniques have been proposed, ranging from classical ML-based methods such as SVM \citep{ML-PCR,Jackson2003} to recent transformer-based models \citep{chalkidis-etal-2019-neural,malik2021ildc}. Researchers have also proposed legal domain-specific language models such as LegalBERT \citep{chalkidis2020legalbert}, CaseLawBERT \citep{zheng2021caselaw} and InLegalBERT and InCaseLawBERT~\citep{paul2023pretrained}. However, legal LLMs have shown limited success and have not demonstrated generalization and transfer learning capabilities \citep{chalkidis2023chatgpt,malik2021ildc,joshi-etal-2023-ucreat}. 

\noindent\textbf{Comparison with Existing Benchmarks:} 
Benchmarks have played a crucial role in the development of better techniques and models in almost every domain, such as computer vision \citep{deng2009imagenet,guo2014benchmark,wu2013online} and reinforcement learning \citep{laskin2021urlb,cobbe2020leveraging,zhang2018natural}. Similarly, in the NLP domain, various benchmarks have been proposed, for example, GLUE \citep{wang2018glue}, Super-GLUE \citep{wang2019superglue}, XTREME \citep{hu2020xtreme}, CLUE \citep{xu2020clue}, GLGE \cite{liu2020glge}, and IndicNLPSuite \citep{kakwani2020indicnlpsuite}. However, these benchmarks focus on the general NLP domain, and models developed for the generic domains do not perform well for the legal domain \cite{malik-rr-2021,joshi-etal-2023-ucreat}. Similar attempts have thus been made for the legal domain; for example, \citet{chalkidis-etal-2022-lexglue} developed \text{LexGLUE}, a specialized English language benchmark (restricted to EU and US legal systems) for evaluating legal NLP models, by consolidating existing datasets for various tasks. 
LexGLUE introduces six main (all classification-based) tasks: violated article identification, case issue classification, concept identification, contract topic prediction, unfair contractual terms identification, and case holding identification. 
\citet{niklaus2023lextreme} have proposed \text{LEXTREME}, a multi-lingual (24 EU languages) legal NLP benchmark (all tasks classification-based) restricted to EU and Brazilian jurisdictions. \citet{chalkidis2022fairlex} have introduced \text{FAIRLEX}, a multi-lingual benchmark consisting of cases from 5 languages and 4 jurisdictions, to test the fairness of different models on legal judgment and topic prediction. \citet{hwang2022multi} have introduced \text{LBOX} benchmark for the Korean legal system. The benchmark targets tasks related to classification and summarization; the documents are in Korean. Recently, \citet{guha2023legalbench} released \text{LegalBench}, a large, collaborative legal benchmark (restricted to US legal system) consisting of 162 tasks (in English) to test the reasoning abilities of LLMs. The tasks belong to six different categories of legal reasoning and address various stages in the pipeline of the litigation process. LegalBench is primarily focused on testing the ability of LLMs to handle legal processes at various stages of litigation; consequently, the tasks involve shorter texts (avg. length $\sim$ 200 words). To benchmark LLMs for Chinese law, \citet{fei2023lawbench} released LawBENCH, a benchmark consisting of 20 tasks (in Chinese) to evaluate the capability of LLMs to memorize and understand legal knowledge. Most of these tasks consist of longer texts than LegalBench (avg. length $\sim$ 300 words).

\begin{table}
    \tiny
    \centering
    \renewcommand{\arraystretch}{1.1}
    \begin{tabular}
    {lp{0.15\columnwidth}p{0.13\columnwidth}p{0.15\columnwidth}p{0.10\columnwidth}}
        \toprule
         \textbf{Dataset} &  \textbf{Jurisdictions} &  \textbf{System} &  \textbf{Task types} & \textbf{Languages}\\ \toprule
         {LexGLUE}&  U.S., E.U.&  Predominantly  Civil Law&  Classification& English\\
         {LEXTREME}&  E.U., Brazil&  Predominantly Civil Law&  Classification& E.U.\\
         {FAIRLEX}&  E.U., U.S., China, Switzerland&  Predominantly Civil Law&  Fairness evaluation & E.U., Chinese\\
         {LBOX}&  Korea&  Civil Law&  Classification, Generation& Korean\\
         {LEGALBENCH}&  Multiple&  Common \& Civil Law&  Generation& English\\
         {LAWBENCH}&  China &  Civil Law&  Classification, Generation, Extraction& Chinese\\ \midrule
         {IL-TUR (ours)}&  India&  Common \& Civil Law&  Classification, Retrieval, Generation, Extraction & English, Indian\\ \bottomrule
    \end{tabular}
    \caption{Comparison of different L-NLP benchmarks.}
    \label{tab:legal-benchmarks}
    \vspace{-5mm}
\end{table}

\begin{table*}[t]
    \centering
    \small
    \renewcommand{\arraystretch}{0.8}
    \begin{tabular}{l c c c m{0.46\textwidth}}
       \toprule
        \textbf{Task} & \begin{tabular}{@{}c@{}} \textbf{Dataset} \\ \textbf{(Language)} \end{tabular} & \begin{tabular}{@{}c@{}} \textbf{Avg.} \\ \textbf{\#Words} \end{tabular} & \begin{tabular}{@{}c@{}} \textbf{Task} \\ \textbf{Type} \end{tabular} & \multicolumn{1}{c}{\textbf{Key Skills Required}} \\ \toprule
        \texttt{L-NER} & \begin{tabular}{@{}c@{}} 105 docs \\ 650k words \\ (English) \end{tabular} & 6,180 &\begin{tabular}{@{}c@{}} Sequence \\ Classification \end{tabular} & Foundational task, legal understanding \\ \midrule
         \texttt{RR} & \begin{tabular}{@{}c@{}} 21,184 sentences \\ (English) \end{tabular}& 25,796 & \begin{tabular}{@{}c@{}} Multi-Class \\ Classification \end{tabular} & Foundational task, legal knowledge and legal semantics understanding \\ \midrule
         \texttt{CJPE} & \begin{tabular}{@{}c@{}} \texttt{ILDC} \\ 34k Docs \\ (English) \end{tabular} & 3,336 & \begin{tabular}{@{}c@{}} Classification, \\ Extraction \end{tabular} & Legal understanding and reasoning  \\ \midrule
         \texttt{BAIL} & \begin{tabular}{@{}c@{}} \texttt{HLDC} \\ 176k Docs \\ (Hindi) \end{tabular} & 86 &Classification & Legal understanding (in Hindi) and reasoning  \\ \midrule
         \texttt{LSI} & \begin{tabular}{@{}c@{}} \texttt{ILSI} \\ 65k samples \\ (English) \end{tabular}  & 2,406 &\begin{tabular}{@{}c@{}} Multi-Label \\ Classification \end{tabular} & Understanding of the statutes and their applicability in various factual situations, commonsense knowledge and reasoning \\ \midrule
         \texttt{PCR} & \begin{tabular}{@{}c@{}} \texttt{IL-PCR} \\ 7,070 Docs \\ (English) \end{tabular} & 8,096 &Retrieval & Understanding of facts (commonsense + legal knowledge) and statutes, concept of legal relevance \\ \midrule
         \texttt{SUMM} & \begin{tabular}{@{}c@{}} \texttt{In-Abs} \\ 7,130 Docs \\ (English) \end{tabular} & 4,376 &Generation & Legal understanding and generation  \\ \midrule
         \texttt{L-MT} & \begin{tabular}{@{}c@{}} \texttt{MILPaC} \\ 17,853 text pairs \\ (English \\ and Indian Langs.) \end{tabular} & 49 & Generation & Parallel understanding of the legal text in English and 9 Indian languages  \\
         \bottomrule
    \end{tabular}
    \caption{Summary of Tasks introduced in \benchmarkName.} 
    \label{tab:tasks-summary}
    \vspace{-3mm}
\end{table*}

\noindent \textbf{\benchmarkName\ differs from the existing benchmarks (see Table~\ref{tab:legal-benchmarks})}. First, \benchmarkName\ focuses on multiple tasks that are not restricted to classification but also involve information retrieval, generation, and explanation. Second, via \benchmarkName, we introduce tasks that are grounded in the actual legal workflow and, consequently, are more complex and involve actual long legal documents (average length $4000$ words). In contrast to some of the popular benchmarks, \benchmarkName\ is not introduced to test the law understanding capability of LLMs but rather to address the problems plaguing the judiciary. In the future, if LLMs are replaced by some other class of machine learning models, \benchmarkName\ would still be relevant. In fact, as shown in our experiments, we observe that long legal documents are challenging for LLMs. Third, \benchmarkName\ is based on Indian legal documents. Given that India is the most populous country in the world (population of $\sim 1.4$  billion \citep{un-report-population}) and there is a backlog of almost $43$ million cases, it is imperative to develop benchmarks and datasets for the Indian legal system. 
From the language perspective, \benchmarkName{} benchmark covers English and $9$ major Indian languages. Although \benchmarkName\  is India-specific, the models developed for \benchmarkName\  could provide inspiration (or possibly adapted) for developing models for the legal systems of other countries. Lastly and most importantly, \benchmarkName{} covers tasks related to the common-law system as well as the civil law system. India has a predominantly common-law system, which implies that a judge in a higher court can overrule existing precedents, so the decision may not always be as per the rule book (written statutes and laws). It introduces some subjectivity into the decision-making process and must be backed by solid reasoning, making the tasks in \benchmarkName\ much more difficult. Additionally, India has a civil law system for certain matters (e.g., banking and insurance). In the proposed benchmark, we cover both settings. Moreover, the legal domain has various areas (following common or civil systems) of laws such as criminal, civil, and banking; via the benchmark, we want to test the cross-area generalization capabilities of the models, i.e., how well the models developed on data from one area generalize across other areas. In contrast, Korea, China, (and, to a large extent, the EU) mainly follow civil law where a decision is as per the rule book. \benchmarkName\ aims to fill the voids in the Legal NLP for the Indian setting by introducing some of the foundational tasks that can be useful for various legal applications. 

\section{\benchmarkName: Legal-NLP Benchmark} \label{sec:benchmark} 
Table \ref{tab:tasks-summary} summarizes various tasks proposed in \benchmarkName. The tasks cover multiple aspects of the legal domain and require specialized skills and knowledge to solve them.  

\vspace{-2.3mm}
\subsection{Design Philosophy}\label{sec:philosophy}


We want to develop technology that enables automated semantic and legal understanding of legal documents and processes. We created \benchmarkName\ with the following principles in mind.

\noindent\textbf{1)~Legal Understanding and World Knowledge:} The tasks should cater exclusively to the legal domain. Solving a task should require in-depth knowledge and understanding of the law and its associated areas. Further, the tasks should not be restricted to only classification but should also involve retrieval, generation, and explanation. The proposed tasks address the pain points of processing legal texts (\S\ref{sec:intro}). Moreover, solving legal tasks should require knowledge about the law as well as commonsense knowledge and societal norms about the world (e.g., facts in conjunction with socio-economic conditions in a particular case).
\textbf{2)~Difficulty Level:} The difficulty level should be such that these are not solvable by a layperson (having minimal knowledge and expertise in legal matters). It ensures that general language learners cannot easily solve the tasks, and the tasks would be sufficiently challenging for the current state-of-the-art models (e.g., LLMs). 
\textbf{3)~Language:} Since India is a multi-lingual society, the tasks should cater to the most frequent languages used in the courts. We cover tasks in English and $9$ other Indian languages. 
\textbf{4)~Evaluation:} The tasks should be automatically evaluable, and the metrics used should align with human judgments. 
\textbf{5)~Public Availability:} The data used for the tasks should be publicly available so anyone can use it for research purposes without licensing or copyright restrictions. Further, a leaderboard should be available to compare different systems and models. We release the data via a Creative Common Attribution-NonCommercial-ShareAlike (CC BY-NC-SA) license and create a public leaderboard.

\vspace{-2mm}
\subsection{\benchmarkName\  Tasks} 
\vspace{-1mm}
Based on the design philosophy, in this version of \benchmarkName, we selected eight different tasks. Table \ref{tab:tasks-summary} provides a summary of the tasks. We briefly describe the tasks here; details about the dataset and evaluation metrics are provided in App.~\ref{app:task-details}. 

\begin{itemize}[noitemsep,nosep]
\item \textbf{Legal Named Entity Recognition {(L-NER)}:} This is a newly created task in \benchmarkName. Formally, given a legal document, the task of Legal Named Entity Recognition is to identify entities (set of 12 entity types), namely, Appellant, Respondent, Judge, Appellant Counsel, Respondent Counsel, Court, Authority, Witness, Statute, Precedent, Date, and Case Number. L-NER is different from the standard NER task; if one were to run a standard NER system on a legal document, the judge, petitioner, and respondent would all be labeled with a ``PERSON" tag. Hence, a separate task is needed to identify the legal named entities in the documents. The standard NER (identifying person/organization/location names) can be done by any non-legal professional/person, but identifying the roles of entities involved in a legal case (L-NER) requires an in-depth understanding of the legal terminologies and the law. Hence, we develop a gold-standard dataset for L-NER with the help of law students (details in \ref{app:ner-task}). Moreover, the set of legal entities and corresponding definitions are formulated with the help of legal academicians (experts). 

\item \textbf{Rhetorical Role Prediction (RR):}  As pointed out earlier, legal documents are typically long (avg. length $4000$ words) and highly unstructured, with the legal information spread throughout the document. {Segmenting the long documents into topically coherent units (such as facts, arguments, precedent, statute, etc.) helps highlight the relevant information and reduces human effort. 
These topically coherent units are termed as \textit{Rhetorical Roles} (RR)}. Given a legal document, the task of RR prediction involves assigning RR label(s) to each sentence. We focus on 13 RR labels:  \textit{Fact, Issue, Arguments (Respondent), Argument (Petitioner), Statute, Dissent,  Precedent Relied Upon, Precedent Not Relied Upon, Precedent Overruled, Ruling By Lower Court, Ratio Of The Decision, Ruling By Present Court, None.} Details about RR labels, definitions, and the dataset are provided in the App. \ref{app:rr-task}. 

\item \textbf{Court Judgment Prediction with Explanation (CJPE):} Formally, the task of  Court Judgment Prediction with Explanation (CJPE) involves predicting the final judgment (appeal accepted or denied, i.e., the binary outcome of 0 or 1) for a given judgment document (having facts and other details) and providing the \textit{explanation for the decision}. In this case, the explanations are in the form of the salient sentences that lead to the decision. 
Note that the idea behind this task is \textit{not} to replace human judges but to augment them in decision-making. Furthermore, the task requires the system to explain its decision so that it is interpretable for a human (details in App. \ref{app:cjpe}).  

\item \textbf{Bail Prediction (BAIL):} A large fraction of the pending cases in India are from the district-level courts and have to do with bail applications (\url{https://en.wikipedia.org/wiki/Bail})~\citep{kapoor-etal-2022-hldc}. Many of the district courts in India use Hindi as their official language (also refer to the Limitations section). Given a legal document in the Hindi language (having the facts of the case), the task of Bail Prediction involves predicting if the accused should be granted bail or not (i.e., a binary decision of 0/1) (details in App. \ref{app:bail}). 

\item \textbf{Legal Statue Identification (LSI):} The task of Legal Statute Identification (LSI) is formally defined to automatically identify the relevant statutes given the facts of a case. One of the first steps in the judicial process is finding the applicable statutes/laws based on the facts of the current situation. Manually rummaging through multiple legislation and laws to find out the relevant statutes can be time-consuming, making the LSI task important for reducing the workload and improving efficiency (more details in App. \ref{app:lsi}). 

\item \textbf{Prior Case Retrieval (PCR):} When framing a legal document, legal experts (judges and lawyers) use their expertise to cite previous cases to support their arguments/reasoning. Legal experts have relied on their expertise to cite previous cases; however, with an exponentially growing number of cases, it becomes practically impossible to recall all possible cases. Given a query document (without citations), the task of Prior Case Retrieval (PCR) is to retrieve the legal documents from the candidate pool that are relevant (and hence can be cited) in the given query document (details in App. \ref{app:pcr}). 

\item \textbf{Summarization (SUMM):} Summarization is a standard task in NLP; however, as mentioned in \S \ref{sec:intro}, summarizing legal documents requires legal language understanding and reasoning. The task of summarization involves generating a gist (of a legal document) that captures the critical aspects of the case. 
We focus on abstractive summarization (more details in App. \ref{app:summ}). 

\item \textbf{Legal Machine Translation (L-MT):} In the Indian legal setting, when a case is transferred (due to re-appeal) from a district court to a High court, the corresponding document (typically in a regional language) needs to be translated to English. Additionally, since a large majority of the Indian population is not proficient in English, High Court / Supreme Court documents often need to be translated from English to Indian languages. 
In both scenarios, such translations, if done by humans, become a primary reason for delay in administering justice. Machine translation (MT) can augment human translators who can post-edit the translated document rather than translating from scratch. India is a diverse country with multiple languages across different states; the task of Legal Machine Translation (L-MT) attempts to close the language barrier by encouraging the development of systems for translating legal documents from English to Indian languages and vice-versa. Given that many Indian languages are low-resource, MT becomes even more challenging, requiring specialized models for translating legal documents in low-resource Indian languages. We focus on 9 Indian languages,  namely, Bengali (BN), Hindi (HI), Gujarati (GU), Malayalam (ML), Marathi (MR),
Telugu (TE), Tamil (TA), Punjabi (PA), and Oriya (OR) (details in App. \ref{app:mt}). 
\end{itemize}


\noindent The tasks in \benchmarkName\ require quite varied skills to solve the problem (Table \ref{tab:tasks-summary}). The skills include a deep understanding of language, the ability to generate legal language, foundational knowledge of law and statutes, application of law to social settings (e.g., decision-making in CJPE and BAIL), and the ability to reason using legal principles. The requirement of such a rich set of skills makes \benchmarkName\ quite challenging; a single model struggles to solve all these tasks, as we observed in our experiments with BERT, LegalBERT, InLegalBERT, GPT3.5 and GPT-4 models (\S\ref{sec:models}).


\noindent \textbf{Harmonization of Tasks:} This resource paper introduces a new benchmark for promoting research and development in the Indian legal system. Since it is a benchmark paper, the aim is to bring domain-specific tasks and datasets under one umbrella so that researchers can compare their models across tasks and with respect to each other. Earlier, no such effort was made for the Indian legal NLP domain. Some of the tasks included in the benchmark already exist; however, there is a lack of standardization across these, e.g., each task and dataset follows its file format, evaluation metric, etc. We have collated all these datasets and converted them to a uniform, JSON-based format so that the community can easily understand and use them. We have also collated all the training scripts for these different tasks together and devised a standard evaluation setup for all these tasks. Further, we have created a website (\website) and a public leaderboard that brings all relevant tasks together. The public leaderboard will further promote transparent and fair comparisons of techniques for each task. Moreover, the leaderboard will lead to the development of more sophisticated models (e.g., GLUE~\cite{wang2018glue} and SuperGLUE~\cite{wang2019superglue} benchmarks promoted further research in NLP). Furthermore, to harmonize these tasks, we also conducted experiments with GPT-3.5 and GPT-4 (see \S\ref{sec:models}) on all the tasks (except PCR), which involved converting the data to the desired format for GPT and formulating the prompts and verbalizers. Also, we plan to grow IL-TUR by introducing more new tasks in the future. We would also like to point out that many existing popular NLP benchmarks such as GLUE~\cite{wang2018glue}, SuperGLUE~\cite{wang2019superglue}, as well as legal benchmarks like LEXGLUE~\cite{chalkidis-etal-2022-lexglue} mostly comprised of datasets released by prior works. GLUE and LEXGLUE introduced only one new dataset each, whereas SuperGLUE did not have any new datasets.

\noindent\textbf{Anonymization of datasets:} In order to address ethical concerns (also see Ethical Considerations section) and to prevent the model from developing any bias, we anonymized named entities in the dataset of the relevant tasks, namely RR, CJPE, BAIL, LSI and PCR  (details in App. \ref{app-sec:ethics}). 

\vspace{-2mm}
\subsection{Relevance of Tasks to Litigation Process}
%
In general, considering the pipeline of a litigation process for a case, all the tasks in the IL-TUR benchmark help formulate various ways in which automatic legal language processing can augment legal practitioners. Among the tasks, LSI is considered one of the first steps in the judicial process -- right after identifying the facts, legal personnel must find out the statutes of the law that are violated. Since India follows a mixture of civil and common law systems, identifying the statutes is not the sole basis of legal reasoning; precedent cases must also be considered (PCR task). Subsequently, the final step in the litigation process is to decide the outcome of the case; the CJPE and BAIL tasks are relevant in this case, and human judges can use corresponding models to get suggestions/recommendations. The tasks of L-NER, RR, and SUMM, though not directly required for the judicial process, significantly help legal practitioners (e.g., lawyers conducting legal research to argue an ongoing case) get a quick understanding of the documents. Sometimes, a case gets re-appealed in a higher court, and consequently, the case document (in a regional language) in the lower court needs to be translated into English (L-MT Task).

\section{Models, Experiments and Results} \label{sec:models}

\begin{table}[t]
    \centering
    \small
    \tabcolsep4pt
    \renewcommand{\arraystretch}{1}
    \begin{tabular}{ccccp{0.25\columnwidth}}
       \toprule
        \textbf{Task} & \textbf{SOTA} & \textbf{Metric} & \textbf{Model Details} \\ \midrule
        \texttt{L-NER} & 48.58\%  & strict mF1 & \begin{tabular}{@{}c@{}}InLegalBERT + \\ CRF\end{tabular} \\ \midrule
        
        \texttt{RR} & 69.01\% & mF1 & MTL-BERT \\ \midrule
        \texttt{CJPE} & \begin{tabular}{@{}c@{}} 81.31\%\\ 0.56\\ 0.32 
        \end{tabular} & \begin{tabular}{@{}c@{}} mF1\\ ROUGE-L\\ BLEU \end{tabular}& \begin{tabular}{@{}c@{}}InLegalBERT + \\ BiLSTM\end{tabular} \\ \midrule
        \texttt{BAIL} & 81\% & mF1 & \begin{tabular}{@{}c@{}}TF-IDF + \\ IndicBERT\end{tabular} \\ \midrule
        \texttt{LSI} & 28.08\% & mF1 & \begin{tabular}{@{}c@{}}LeSICiN \\ (Graph-based Model)\end{tabular} \\ \midrule
        \texttt{PCR} & 39.15\% & $\mu$F1@$K$ & Event-Based \\ \midrule
        \texttt{SUMM} & \begin{tabular}{@{}c@{}}0.33\\ 0.86 \end{tabular} & \begin{tabular}{@{}c@{}}ROUGE-L\\ BERTScore \end{tabular} & Legal-LED \\ \midrule
        \texttt{L-MT} & \begin{tabular}{@{}c@{}}0.28\\ 0.32\\ 0.57 \end{tabular} & \begin{tabular}{@{}c@{}}BLEU\\ GLEU\\ chrF++ \end{tabular} & \begin{tabular}{@{}c@{}}MSFT \\ (Microsoft Translation) \end{tabular} \\ 
         \bottomrule
    \end{tabular}
    \caption{Summary of the best result for each task, along with the model that achieved the best result.}
    \label{tab:tasks-results}
    \vspace{-2mm}
\end{table}

\begin{table}[t]
    \centering
    \small
    \setlength{\tabcolsep}{5pt}
    \renewcommand{\arraystretch}{0.5}
    \begin{tabular}{ccccccc}
       \toprule
        \multirow{2}{*}{\textbf{Task}} & \multirow{2}{*}{\textbf{Arch}} & \multicolumn{4}{c}{\textbf{BERT}} \\ \cmidrule(lr){3-6}
        & & \textbf{V} & \textbf{L} & \textbf{InL} & \textbf{Ind} \\ \midrule
        \texttt{L-NER} & Flat(CS) & 39.59\% & 45.58\% & 48.58\% & - \\ \midrule
         \texttt{RR} & Flat(S) & 58\% & 54\% & 58\% & - \\\midrule
         \texttt{CJPE} & Hier & 71.14\% & 78.21\% & 81.31\% & - \\ \midrule
         \texttt{BAIL} & Flat(L) & -  & -  & -  & 76\% \\ \midrule
         \texttt{LSI} &  Hier & 18.44\%  & 21.74\%  & 26.23\% & - \\ \midrule
         \texttt{PCR} & Flat(CS) & 9.24\% & 8.67\% & 7.57\% & - \\ 
         \bottomrule
    \end{tabular}
    \caption[Results of different BERT-based models on tasks of \benchmarkName{}]{Results of different BERT-based models on tasks of \benchmarkName{}. V, L, InL, and Ind refer to Vanilla BERT, LegalBERT, InLegalBERT, and IndicBERT, respectively. All metrics are in terms of macro-F1 (strict mF1 for L-NER). All the BERT-based models are implemented with the default architectures: either in a flat setup by taking individual sentences (S), segmenting long texts (CS), choosing the last 512 tokens for encoding (L), or in a hierarchical (Hier) setup with a BiLSTM on top of BERT.}
    \label{tab:bert-results}
    \vspace{-5mm}
\end{table}

\noindent We extensively experimented with various models for each proposed task, including transformer-based language models. Table \ref{tab:tasks-results} summarizes baseline models and results for all tasks. Due to space limitations, we provide only the top-performing models here; details of experiments (e.g., hyperparameters) and other models are in App.~\ref{app-sec:results}. In general, results indicate that the tasks are far from being solved, and more research is required. 
In particular, we experimented with both generic BERT model~\citep{devlin-etal-2019-bert} and legal domain-specific BERT models: LegalBERT~\citep{chalkidis2020legalbert} (BERT pre-trained on EU legal documents), CaseLawBERT~\citep{zheng2021caselaw} (BERT pre-trained on US legal documents), and InLegalBERT~\citep{paul2023pretrained} (BERT pre-trained on Indian legal documents). For L-NER, InLegalBERT (with CRF on top) shows the best performance, possibly because of in-domain data pre-training. For the RR task, vanilla BERT (or other transformers) and Legal-BERT do not work well; hence, RR prediction is posed as a sequence prediction problem (at the sentence level), and the Multi-Task Learning (MTL) model based on BERT developed by \citet{malik-rr-2021} shows the best performance. 
Since legal documents are long, and BERT has a limitation of 512 tokens in the input, for the CJPE task, hierarchical InLegalBERT (InLegalBERT and BiLSTM on top of that) \citep{paul2023pretrained} works best. For explanations, we use the occlusion method for finding the sentences leading to the final decision \cite{malik2021ildc}. But these fall short of expert-annotated important sentences in terms of ROUGE-L and BLEU scores.
For BAIL prediction, since the documents are in Hindi, IndicBERT \citep{kakwani2020indicnlpsuite}, a BERT model trained on Indian languages, was used. A pre-filtering of salient sentences, followed by IndicBERT, works best \cite{kapoor-etal-2022-hldc}. 
For the LSI task, we conduct experiments with hierarchical LegalBERT and InLegalBERT, along with LeSICiN, a graph-based method proposed by \citet{paul2022lsi}. We observe that LeSICIN outperforms the BERT-based methods.
For the PCR task, an event-based model works the best \citep{joshi-etal-2023-ucreat}. An event refers to an action/activity (in the form of a predicate (typically a verb) and corresponding arguments) mentioned in the document.
For SUMM, Legal-LED \citep{legal-led} performs the best, and the commercially available Microsoft Azure Cognitive Services Translation API works best for the L-MT task. 
In general, across all tasks except LSI, PCR, SUMM, and L-MT, BERT (or its variant) performs the best. In order to compare the same model type across all tasks, we also experimented with BERT and its variants across all tasks that do not require text generation (i.e., SUMM and L-MT). Specifically, we took BERT (\texttt{bert-base-uncased}), LegalBERT, and InLegalBERT as the encoders and ran it either in the flat text setup (either over the last 512 tokens or individual sentences/chunks) or hierarchical setup (full document), as per the task requirement. For BAIL, we use IndicBERT since the documents are in Hindi. These results are reported in Table~\ref{tab:bert-results}. In general, except for PCR, we observe that performance increases going from BERT to LegalBERT to InLegalBERT, which correlates with the degree of in-domain pre-training.


\begin{table}[t]
    \small
    \centering
    \setlength{\tabcolsep}{3pt}
    \begin{tabular}{cccccc}
        \toprule
         \multirow{2}{*}{\textbf{Model}}&  \multirow{2}{*}{\textbf{Trained On}}&  \multicolumn{3}{c}{\textbf{ROUGE}}& \multirow{2}{*}{\textbf{BERTScore}}\\ \cmidrule(lr){3-5}
         &  &  \textbf{R-1}&  \textbf{R-2}&  \textbf{R-L}& \\ \midrule
         \multirow{2}{*}{\begin{tabular}{@{}c@{}}Summa- \\ RuNNer \\ \end{tabular}}&  In-Abs&  0.604&  0.264&  0.225& 0.828\\ 
         &  UK-Abs&  0.493&  0.255&  0.274& 0.849\\ \midrule
         \multirow{2}{*}{\begin{tabular}{@{}c@{}}Legal \\ LED \\ \end{tabular}}&  In-Abs&  0.557&  0.244&  0.242& 0.844\\
         &  UK-Abs&  0.482&  0.186&  0.264& 0.851\\ \midrule
    \end{tabular}
    \caption{Performance of SummaRuNNer and Legal LED on UK-Abs test set after being trained on In-Abs~(part of \benchmarkName) and UK-Abs train sets.}
    \label{tab:uk-abs-results}
    \vspace{-4mm}
\end{table}

\noindent\textbf{Model generalization beyond Indian jurisdiction:} Law is country/region specific. The laws of one country cannot be directly applied to another country. Hence, the legal NLP models developed for one region are less likely to generalize across countries. A similar pattern is also observed among human lawyers, i.e., an Indian lawyer cannot practice directly in EU/US jurisdictions. Moreover, even many of the tasks are jurisdiction-specific, e.g., tasks like BAIL, CJPE (since the processes used for deciding bail are different across countries), and LSI (since statutes are country-specific by nature), PCR, and L-MT require a deep understanding of the Indian legal system. 
For tasks like L-NER and RR, one could test the generalization capabilities of models across jurisdictions; however, we could not find equivalent datasets (with the same labels) in other jurisdictions. For the summarization task, one can easily check the generalization capability of models across legal systems. So we conducted cross-jurisdiction experiments on abstractive summarization of \textit{UK Supreme Court documents} using the UK-Abs dataset~\citep{shukla2022summ}. 
Uk-Abs consists of gold standard summaries released by the UK Supreme Court as press summaries. We experimented with two best-performing models: SummaRuNNer and Legal LED (trained on the In-Abs dataset, which is part of \benchmarkName{}). These are used to generate summaries on the test set of UK-Abs (100 documents). 
Results are reported in Table~\ref{tab:uk-abs-results}. For comparison, we also report the results of these models on the UK-Abs test set when trained on the train set of UK-Abs itself. The results show that the summarization models trained on In-Abs perform decently when tested on UK-Abs (in a zero-shot setting). Both SummaRuNNer and Legal LED trained on In-Abs outperform their counterparts trained over UK-Abs in terms of ROUGE-1 and ROUGE-2 and achieve comparable ROUGE-L and BERT Scores. This experiment further shows the utility of our \benchmarkName{} benchmark. Nevertheless, the generalization of models across jurisdictions requires more research in cross-jurisdiction domain adaptation techniques; we leave this for future work.

\begin{table*}[t]
    \centering
    \small
    \renewcommand{\arraystretch}{1}
    \begin{tabular}{ccccccccc}
       \toprule
        \multirow{2}{*}{\textbf{Task}} & \multicolumn{3}{c}{\textbf{GPT-3.5}} & \multicolumn{3}{c}{\textbf{GPT-4}} & \multirow{2}{*}{\textbf{SOTA}} & \multirow{2}{*}{\textbf{Metric}} \\
        \cmidrule(lr){2-4} \cmidrule(lr){5-7}
        & \textbf{0-Shot} & \textbf{1-Shot} & \textbf{2-Shot} & \textbf{0-Shot} & \textbf{1-Shot} & \textbf{2-Shot} & & \\ \midrule
        
        \texttt{L-NER} & 30.59\% & 23.68\% & \uline{32.84}\%* & 13.65\% & 10.51\% & 24.03\% & \textbf{48.58}\% & strict mF1  \\ \midrule
         \texttt{RR} & 30.95\%  & 30.05\%  & 30.31\% & 37.37\% & 37.43\% & \underline{38.18}\% & \textbf{69.01}\% & mF1 \\ \midrule
         \texttt{CJPE} & \begin{tabular}{@{}c@{}}54.17\%\\ 0.30\\ 0.08\\ \end{tabular} & \begin{tabular}{@{}c@{}} 51.46\%\\ 0.29 \\ 0.15 \end{tabular} & \begin{tabular}{@{}c@{}} 56.74\% \\0.30 \\0.113 \end{tabular} & \begin{tabular}{@{}c@{}} \underline{68.29}\% \\0.40 \\0.14 \end{tabular} & \begin{tabular}{@{}c@{}} 47.26\% \\0.39 \\0.16 \end{tabular} & \begin{tabular}{@{}c@{}} 60.44\% \\\uline{0.43} \\\uline{0.18} \end{tabular} & \begin{tabular}{@{}c@{}} \textbf{81.31}\% \\ \textbf{0.56} \\ \textbf{0.32} \end{tabular} & \begin{tabular}{@{}c@{}} mF1 \\ ROUGE-L \\ BLEU \end{tabular} \\ \midrule
         \texttt{BAIL} &  51.04\%  & 46.35\%  & 61.0\% & 51.46\% & 56.90\% & \underline{66.67}\% & \textbf{81}\% & mF1 \\ \midrule
         \texttt{LSI} &  21.55\%  & 22.61\%  & 21.40\% & \uline{23.99} & 22.26 & 20.53 & \textbf{28.08}\% & mF1  \\ \midrule
         \texttt{SUMM} &  \begin{tabular}{@{}c@{}} 0.21 \\ \underline{0.85} \end{tabular} & \begin{tabular}{@{}c@{}} 0.20 \\ 0.84 \end{tabular} & \begin{tabular}{@{}c@{}}0.22 \\ 0.84 \end{tabular} & \begin{tabular}{@{}c@{}} \uline{0.23} \\ \uline{0.85} \end{tabular} & \begin{tabular}{@{}c@{}}0.16 \\ 0.81 \end{tabular} & \begin{tabular}{@{}c@{}}0.17 \\ 0.81 \end{tabular} & \begin{tabular}{@{}c@{}} \textbf{0.33} \\ \textbf{0.86} \end{tabular} & \begin{tabular}{@{}c@{}} ROUGE-L\\ BERTScore \end{tabular} \\ \midrule
         \texttt{L-MT} & \begin{tabular}{@{}c@{}} 0.23\\ 0.28 \\ 0.42 \end{tabular} & \begin{tabular}{@{}c@{}}0.25 \\0.28 \\ 0.43 \end{tabular} & \begin{tabular}{@{}c@{}} 0.26 \\ 0.29 \\ 0.43 \end{tabular} & \begin{tabular}{@{}c@{}} 0.33 \\ 0.36 \\ 0.50 \end{tabular} & \begin{tabular}{@{}c@{}} 0.35 \\ 0.38 \\ 0.52 \end{tabular} & \begin{tabular}{@{}c@{}} \uline{\textbf{0.36}} \\ \uline{\textbf{0.39}} \\ \uline{0.53} \end{tabular} &  \begin{tabular}{@{}c@{}} 0.28 \\ 0.32 \\ \textbf{0.57} \end{tabular} & \begin{tabular}{@{}c@{}} BLEU \\ GLEU \\ chrF++ \end{tabular} \\ 
         \bottomrule
    \end{tabular}
    \caption[Performance of OpenAI GPT models]{Performance of OpenAI GPT-3.5 (\texttt{gpt-3.5-turbo-16k}) and GPT-4 (\texttt{gpt-4-turbo}) model on various tasks for zero-shot, one-shot and two-shot settings. The SOTA corresponds to the best-performing model as given in Table \ref{tab:tasks-results}. The best result for each task is marked in \textbf{boldface}. The best GPT-based result for each task is \uline{underlined}.} 
    \label{tab:llm-results}
    \vspace{-3mm}
\end{table*}

\noindent\textbf{Experiments with LLMs:} We also conducted experiments with LLMs. In particular, we experimented with large models (in terms of the number of parameters) like Open-AI GPT-3.5 (\texttt{gpt-3.5-turbo-16k}) and GPT-4 (\texttt{gpt-4-turbo}) 
and smaller models like GPT-Neo~\cite{gpt-neo} family of three models (GPT-Neo-125M, GPT-Neo-1.3B, GPT-Neo-2.7B), GPT-J-6B~\cite{gpt-j}, Llama-2-7b-chat-hf~\cite{touvron2023llama}, and Mistral-7B-v0.1~\cite{jiang2023mistral}. We experimented with zero-shot settings and In-Context Learning (ICL)-based settings (one-shot and two-shots).  
Table~\ref{tab:llm-results} shows the results for Open-AI GPT-3.5 and GPT-4 models (details about prompts and other settings in App. \ref{app-sec:llms}). We could not experiment with ICL for PCR since it requires a comparison between the query document and the pool of all candidate documents, and passing the content of all the documents to GPT (or other LLMs) exceeds the token length limit even for GPT-4 (having context length of 16,000 tokens). In the future, we would like to experiment with a two-stage retrieval process to feed a subset of candidates to GPT instead of all of them. As observed, the GPT models perform worse than the SOTA models for each task, except for GPT-4 on MiLPAC for L-MT. It may be because the tasks are quite complex and require reasoning across long contexts. Also, for some tasks like L-NER and RR, it can be hard to come up with output formats that the models can understand in a zero-shot setting. Results for one-shot and two-shot show a similar trend in most cases. In some cases, one-shot performance is worse than zero-shot performance (also observed in other works \cite{brown2020language-incontext-learning}), while ICL has no effect on some tasks. Overall, in most cases and under most settings, GPT-4 outperforms GPT-3.5 by significant margins. Experiments with smaller models (GPT-Neo-125M, GPT-Neo-1.3B, GPT-Neo-2.7B, GPT-J-6B, Llama-2-7b-chat-hf, and Mistral-7B-v0.1) showed similar trends (details in App. \ref{app-sec:llms}). 

\noindent\textbf{Discussion:} Tasks in \benchmarkName\ are quite varied, requiring different types of knowledge and skills. Developing systems for the legal domain is not easy due to the inherent challenges (\S\ref{sec:intro}). Moreover, legal datasets are expensive to annotate; consequently, annotated legal datasets are relatively small in size, and hence, learning in a low-resource setting is challenging. Experiments indicate that transformers fine-tuned on legal texts have shown limited success in the legal domain. Further, LLMs like Chat-GPT, which have demonstrated SOTA results in other domains (and have been shown to pass the bar exam~\citep{chalkidis2023chatgpt}), have not performed well on the IL-TUR benchmark, indicative of further research required in the legal domain.

\vspace{-1mm}
\section{Conclusion and Future Directions} \label{sec:conclusion}
\vspace{-1mm}

This paper presented \benchmarkName, a benchmark for Indian Legal Text Understanding and Reasoning. The benchmark has eight tasks requiring different types of legal skills to solve. Results indicate that the tasks are far from solved using state-of-the-art transformer-based models and LLMs. The list of tasks in \benchmarkName\ is not exhaustive, and we plan to expand the list of tasks in the future; for example, we are working on developing foundational tasks like \textbf{Legal Coreference Resolution (L-Coref)} that are required for various applications such as information extraction and knowledge graph creation. Although such tasks have been addressed well in general NLP, our initial experiments show that using SOTA transformer models (which have become part of standard NLP toolkits) do not perform well on legal texts. Due to the usage of specialized terms, new models are needed for the legal domain. On the modeling side, in the future, we plan to develop one model that generalizes and works across all the tasks (e.g., mT5 \citep{xue2020mt5} and Multi-task Adapters \citep{pfeiffer2020mad}). Overall, we hope that \benchmarkName\ (along with its leaderboard) and its successive versions would create excitement in the Legal-NLP community and lead to the development of new technologies that could benefit society immensely and facilitate fair access to justice, a fundamental human right.  

\section*{Limitations} \label{sec:limitations}

\benchmarkName\ is a first step towards creating a benchmark for the Indian legal domain, which desperately needs technological solutions. The benchmark is not perfect and has certain limitations. Given the dynamic nature of the legal domain, new cases and precedents keep getting added. Hence, we plan to keep updating \benchmarkName\ in the future. The legal domain is vast and covers various areas such as criminal law, civil law, banking, insurance, etc. In \benchmarkName, we could not cover each of the sub-domains in each task as it is a time-consuming and expensive affair to annotate many documents. One of our goals for \benchmarkName\ is to test the cross-area generalization abilities of models; nevertheless, we would expand the datasets of each task in the future. \benchmarkName\ is multi-lingual only concerning the L-MT task. Additionally, the BAIL task is in Hindi. All the High Courts and the Supreme Court in India use English as the official language. Hindi is the prominent language used in the district courts in most north Indian states. Nevertheless, India is a multi-lingual society, and legal models for other languages should also be developed for more tasks in the legal domain. We plan to extend the benchmark in the future and include some more tasks in Indian languages. The main challenge in doing so is a scarcity of legal data in regional languages in digitized formats from lower courts. Datasets of some of the tasks (e.g., LSI) use ML-based models (that may not be perfect) in the dataset creation process (e.g., fact extraction in the case of LSI). Extracting facts manually at a large scale is an expensive and time-consuming effort; in the future, we plan to employ legal professionals and create a more refined dataset. Regarding explainability, at present, we mainly address model explainability in the context of the CJPE task. For discussion regarding other tasks, please refer to App.~\ref{app-sec:limitations}. Regarding LLM experiments, some of the tasks, such as BAIL and CJPE, require the entire document to be a part of the model's input. Obtaining LLM predictions overall test set samples is challenging in terms of expense and computation. Hence, we evaluated over a small subset, assuming that it is a good proxy of LLM performance. 
Lastly, the benchmark has only eight tasks. Creating legal tasks is time-consuming and expensive since it requires the help of legal experts. Nevertheless, as explained earlier, \benchmarkName\ is a work in progress, and we will keep growing by adding more tasks. In this work, we presented different models for various tasks; although many of the models (e.g., BERT, GPT) are common across all tasks, in the future, we plan to develop a single model that could solve all the tasks (e.g., mT5) with reasonable accuracy.

\vspace{-2mm}
\section*{Ethical Considerations} \label{sec:ethics}
\vspace{-3mm}

We use publicly available and open-source datasets for the tasks; no copyright is infringed. To the best of our knowledge, five of the proposed tasks (L-NER, RR, LSI, PCR, and Summ) do not have any direct ethical consequences since the proposed tasks are mainly related to information retrieval and summarization. Moreover, the tasks are meant to encourage the development of systems that would lead to streamlining the legal workflow and will not directly affect the life of any personnel. 

\noindent For the LSI task, to prevent any bias in the model, named entities in the dataset were anonymized (details in App.~\ref{app-sec:ethics}). Similarly, the named entities were anonymized in the RR and PCR datasets. 
App.~\ref{app-sec:ethics} provides more details about various measures and potential risks associated with failure to anonymize legal data.
The documents are selected randomly for all tasks to avoid bias towards any entity, organization, or law. 

\noindent Two tasks (CJPE and BAIL) have ethical considerations. Given a large quantum of pending cases in Indian courts, these tasks aim to develop systems that augment judges and \textit{not} replace them; consequently, the systems are meant to provide recommendations, and a human judge takes the final decision. 
We follow all the steps as done by \citet{malik2021ildc,kapoor-etal-2022-hldc} to avoid any bias in the data for these two tasks. For example, we removed cases (documents) related to sensitive issues like rape and sexual violence, and named entities were anonymized.

\noindent \textbf{Note that we do not endorse the use of the benchmark data for non-research (commercial and real-life) applications, and the primary motivation for creating the \benchmarkName\ benchmark is to consolidate all the research happening in parallel for the Indian Legal domain.} Hence, we will release the benchmark and datasets under the Creative Common Attribution-NonCommercial-Share-Alike (CC BY-NC-SA) license. Moreover, we believe providing a platform by maintaining a common leaderboard for multiple tasks will advance the field with more transparency and reproducibility. 




\section*{Acknowledgements}
The work is partially supported by the Technology Innovation Hub (TIH) on AI for Interdisciplinary Cyber-Physical Systems (AI4ICPS) at IIT Kharagpur, set up under the aegis of DST, Government of India (GoI), through the project ``NyayKosh: Multilingual Resources for AI-based Legal Analytics''. Abhinav Joshi and Shounak Paul are supported by the Prime Minister’s Research Fellowship from the Government of India.

\bibliography{references}




\clearpage
\newpage

\appendix
\section*{Appendix}

\appendix


\titlecontents{section}[18pt]{\vspace{0.05em}}{\contentslabel{1.5em}}{}
{\titlerule*[0.5pc]{.}\contentspage} 

\titlecontents{table}[0pt]{\vspace{0.05em}}{\contentslabel{1em}}{}
{\titlerule*[0.5pc]{.}\contentspage} 

\startcontents[appendix] 
\section*{Table of Contents} 
\printcontents[appendix]{section}{0}{\setcounter{tocdepth}{4}} 

\startlist[appendix]{lot} 
\section*{List of Tables} 
\printlist[appendix]{lot}{}{\setcounter{tocdepth}{1}} 

\startlist[appendix]{lof} 
\section*{List of Figures} 
\printlist[appendix]{lof}{}{\setcounter{tocdepth}{1}} 

\newpage

\section{Tasks and Dataset Details} \label{app:task-details}

We will release the baseline codes along with a compiled list of task-specific datasets and evaluation scripts with the camera-ready version of the paper. The consolidated leaderboard website for the benchmark will be made public with the camera-ready release. We now describe the tasks in the benchmark in detail. 




\subsection{Legal Named Entity Recognition (L-NER)}\label{app:ner-task}

\begin{figure}
  \centering
 \includegraphics[width=\linewidth]{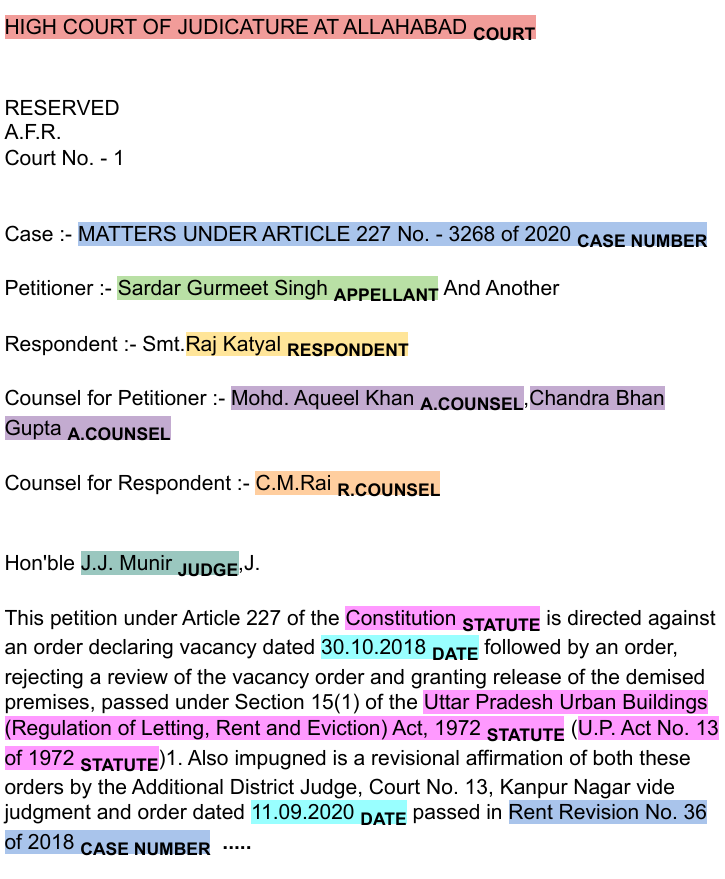}
  \caption[L-NER Dataset example]{Example of L-NER}
  \label{app-fig:ner}
\end{figure}

\noindent\textbf{Task Motivation and Description:} Named Entity Recognition (NER) is a foundational task in NLP~\citep{yadav2019survey}. However, in the legal domain, the types of named entities one may be interested in differ (e.g., judge, petitioner (appellant), and respondent), which may not be identified by a standard NER system. Hence, a separate task is needed to identify the legal named entities in the documents. Note that L-NER is very different from the standard NER; the standard NER (identifying person/organization/location names) requires a language understanding; in contrast, identifying the roles of entities involved in a legal case (L-NER) requires an understanding of the legal terminologies. Hence, we develop a gold-standard dataset for L-NER annotated with the help of law students (details in App. \ref{app:task-details}). Moreover, the set of legal entities and corresponding definitions are formulated with the help of legal academicians (experts). \textbf{Formally, given a legal document, the task of Legal Named Entity Recognition is to identify entities (set of 12 entity types), namely, Appellant, Respondent, Judge, Appellant Counsel, Respondent Counsel, Court, Authority, Witness, Statute, Precedent, Date, and Case Number.} Fig. \ref{app-fig:ner} shows an example. Table \ref{tab:ner-classes} shows the definition for 12 NE types/classes.   

\noindent\textbf{Dataset:} We collected a total of $105$ case documents in English (a total of $650$K words and $12.5$K entities). Table~\ref{tab:ner-stats} lists some important statistics about the NER dataset. The NE type label statistics are displayed along with the class descriptions in Table~\ref{tab:ner-classes}. 


\begin{table*}[!h]
\centering
\small
\renewcommand{\arraystretch}{1.3}
\begin{tabular}{l l l p{0.4\textwidth}} 
\toprule
\textbf{Broad Category} & \textbf{Label} & \textbf{Frequency} & \textbf{Description} \\ \midrule 
\multirow{2}{*}{Party} & APPELLANT (APP) & 660 & Party filing an appeal to the court \\ 
 & RESPONDENT (RESP) & 516 & Party against whom appeal has been filed \\ 
\midrule
\multirow{3}{*}{Legal Professional} & JUDGE (JUD) & 366 & \begin{tabular}[c]{@{}l@{}}Judge of the current or prior/cited cases \end{tabular}  \\ 
 & A.COUNSEL (AC) & 288 & Lawyer(s) on behalf of the appellant(s) \\ 
 & R.COUNSEL (RC) & 255 & Lawyer(s) on behalf of the respondent(s) \\ 
\midrule
\multirow{2}{*}{Organizations} & COURT (CRT) & 1,572 & Any court occurring in the document \\ 
 & AUTHORITY (AUTH) & 1,342 & Any organization/body having administrative/legal authority \\ 
\midrule
Other Person(s) & WITNESS (WIT) & 312 & Witness(es) who are testifying in the case \\ 
\midrule
\multirow{2}{*}{Legal References} & STATUTE (STAT) & 2,055 & Citation to legal acts \\ 
 & PRECEDENT (PREC) & 1,804 & Citation to prior cases \\ 
\midrule
\multirow{2}{*}{\begin{tabular}[c]{@{}l@{}}Legal Artefacts\end{tabular}} & DATE & 2,316 & Mention of any date in the case \\ 
& CASE NO. (CN)  & 1,102 & Mention of any case number, including that of the current case \\ \bottomrule
\end{tabular}
\caption[L-NER Dataset labels]{Named Entity (NE) types used in the L-NER dataset}
\label{tab:ner-classes}
\end{table*}


\begin{table}[htbp]
\small
\renewcommand{\arraystretch}{1.3}
\setlength\tabcolsep{15pt}
\hspace{0.2cm}
\centering
\begin{tabular}{p{0.6\columnwidth}c}
\toprule
\# Documents & 105 \\
\# Labels & 12 \\ \midrule
Total no. of words & 648,937 \\
Avg. Document Size (in \#words) & 6180.35 \\ \midrule
Total no. of entities (All occurrences) & 12,588 \\
Total no. of entities (Unique occ.) & 5,658 \\
Avg. no. of entities per doc (All occ.) & 119.89 \\
Avg. no. of entities per doc (Unique occ.) & 53.89 \\
\bottomrule
\end{tabular}
\caption[L-NER Dataset Statistics]{The dataset statistics for the L-NER task}
\label{tab:ner-stats}
\end{table}

\noindent\textbf{Annotation Details:} For the L-NER task, we collected a total of 105 cases publicly available from the Supreme Court and a few High Courts of India by scrapping the website: \url{https://www.indiankanoon.org}. Please note that the IndianKanoon website allows free downloads of public documents. In discussion with legal experts, we decided on a comprehensive set of 12 NE (Named Entity) classes suited for the legal domain (Table~\ref{tab:ner-classes}). Two law students from a reputed law college in India were tasked with annotating the case documents. The annotation procedure involved the following steps:

\begin{itemize}[noitemsep,nosep]
    \item To ensure that entity spans are marked consistently, we discussed with both annotators how to mark every label. Such decisions involved leaving out prefixes/salutations such as `Shri' (a polite way to address Mr. in the Indian context) and `Smt.' (a polite way to address Ms. in the Indian context), `Justice' (Honorific for a Judge), etc., from the entity names, including the (optional) precedent citations that follow case titles as part of the precedent (PREC) entities, and so on.
    \item We randomly chose a set of 25 documents, and each annotator worked on all 25 documents independently based on the rules devised in the previous step.
    \item We observed a high degree of agreement between the annotators for these 25 documents (Cohen's Kappa: 0.82, Krippendorff's Alpha: 0.85).
    \item Both annotators worked together to resolve the disagreements to arrive at one single consolidated set of annotations for these 25 documents.
    \item Above steps performed over 25 documents calibrated the annotators and led to a high degree of agreement among them. Since annotation is an expensive and time-consuming process, the remaining 80 documents were split equally between the two annotators for annotation. 
\end{itemize}


\noindent\textbf{Task Evaluation:} NER can be formulated as a  sequence prediction task, where each word receives either of the labels \{\texttt{B-X}, \texttt{I-X}, \texttt{O}\} as per the popular `B-I-O' scheme \citep{yadav2019survey} (`X' represents any of the legal classes we are interested in).
We use standard metrics of \emph{strict} macro-averaged precision, recall, and F1 score for evaluation. The \emph{strict} score assumes a correct match only if \textit{both} the entity boundary and entity type are correctly predicted. 
L-NER evaluation F1 score metric is computed using \url{https://pypi.org/project/nervaluate/}. We use strict macro-averaged scores in our setup. The \emph{strict} scoring mechanism ensures that a match is considered correct if the entity span and entity type are the same. In other words, if either the span is incorrect (the model predicts more/fewer tokens as part of the entity) or the predicted label type does not match the ground truth, the match is considered incorrect.

\noindent\textbf{Comparison with existing L-NER datasets:} Recently, \citet{kalamkar-etal-2022-named} released a dataset for L-NER over Indian legal documents. However, unlike our dataset, which comprises of \textit{full-length documents} annotated with every occurrence of every NE, the dataset by \citet{kalamkar-etal-2022-named} consists of segments of documents and not full documents. This is a crucial difference since models trained on our data will be able to detect NEs even when provided with a snippet of a case document.
There is also a slight variation in the set of NEs considered in our dataset as compared to those considered by \citet{kalamkar-etal-2022-corpus}, although most common entity types have been covered in both datasets.




\subsection{Rhetorical Role Prediction (RR)} \label{app:rr-task}

\begin{figure*}
  \centering
   \includegraphics[width=\textwidth]{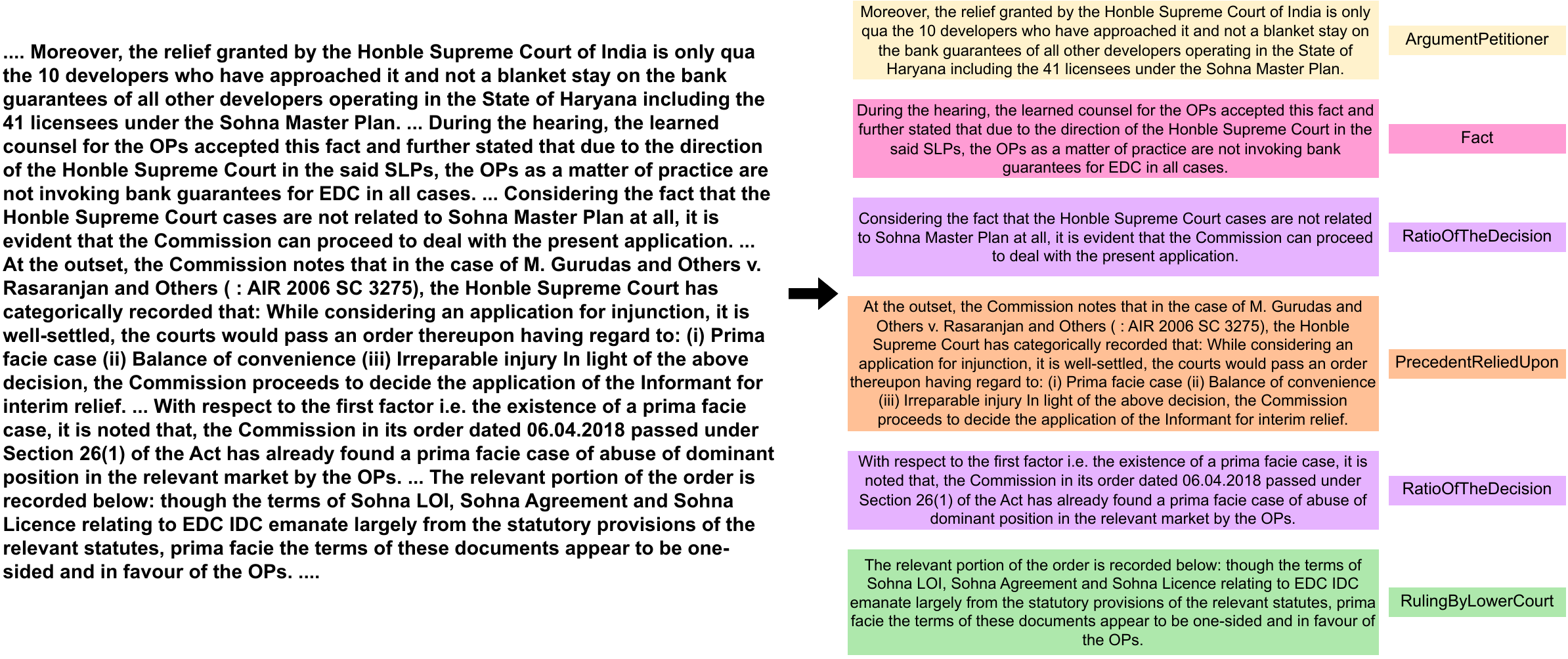}
  \caption[RR Dataset example]{Example of the Rhetorical Role Prediction Task \citep{kalamkar-etal-2022-corpus}}
  \label{app-fig:rr}
\end{figure*}

\noindent\textbf{Task Motivation and Description:} As pointed out earlier, legal documents are typically long (avg. length $4000$ words) and highly unstructured, with legal information spread throughout the document. {Segmenting the long documents into topically coherent units (such as facts, arguments, precedent, statute, etc.) not only helps highlight the relevant information but also reduces human effort when going through a long list of documents. These topically coherent units are termed as \textit{Rhetorical Roles} (RR)}. \textbf{Given a legal document, the task of RR prediction involves assigning RR label(s) to each sentence.} The sentences are annotated with 13 RRs by as many as six legal experts (from a reputed Indian law school). The 13 RR labels are: \textit{Fact, Issue, Arguments (Respondent), Argument (Petitioner), Statute, Dissent,  Precedent Relied Upon, Precedent Not Relied Upon, Precedent Overruled, Ruling By Lower Court, Ratio Of The Decision, Ruling By Present Court, None.} The definition of each RR label is given in Table \ref{tab:rr-definition}. We utilize the dataset and role definitions provided by prior work on structuring Indian legal documents~\citep{malik-rr-2021}. Fig.~\ref{app-fig:rr} shows an excerpt from a legal document annotated with RR labels. RR Prediction is a foundational task that helps structure the information and thus aids downstream applications related to document understanding, information extraction, summarization, and retrieval. 

\begin{figure}[h]
    \centering
    \includegraphics[width=\linewidth]{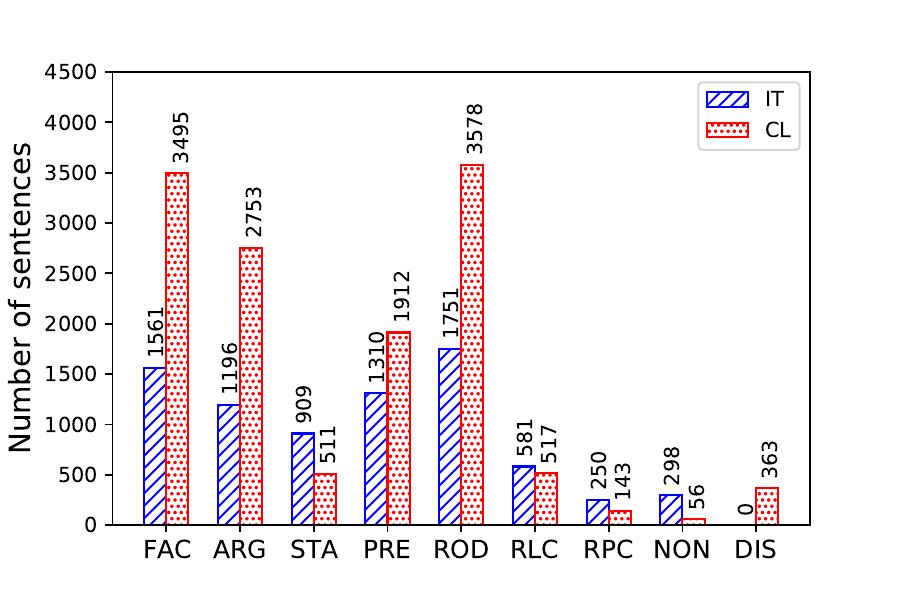}
    \caption[Distribution of RR labels]{Distribution of RR labels in IT and CL documents \citep{malik-rr-2021}.}
    \label{fig:RRstats}
\end{figure}

\noindent\textbf{Dataset:} 
For this task, we use the dataset developed by \citet{malik-rr-2021}  primarily due to the large number of annotations by several Law academicians and public availability.
The dataset consists of $21,184$ sentences from legal documents (in English) about banking and competition law. 
The RR dataset was created by scrapping (from IndianKanoon website: \url{https://indiankanoon.org/}) publicly available documents from the Supreme Court of India, High Courts, and Tribunal courts. The documents pertain to Banking/Income Tax law (IT) and Competition Law (CL) (also called as Anti-Trust Law in the US). The dataset consists of $21,184$ sentences annotated with 13 RRs. Figure \ref{fig:RRstats} shows the distribution of RR labels. The dataset is split randomly (at document level) into 80\% train, 10\% validation, and 10\% test set.  

\noindent\textbf{Annotation Details:} The dataset was annotated by six legal experts (graduate law student researchers), three annotated CL documents, and the remaining three annotated IT documents \citep{malik-rr-2021}. The annotators showed a high degree of agreement. The Fleiss kappa \citep{fleiss2013statistical} between the annotators is 0.65 for the IT domain and 0.87 for the CL domain, indicating a substantial agreement between annotators. Annotating RR is not a trivial task, and annotators can have disagreements.
Several strategies were employed to resolve these disagreements.
More details about annotation case studies can be found in \citet{malik-rr-2021}.  

\noindent\textbf{Evaluation:} RR Prediction is evaluated using standard Macro F1 metric. Macro F1 is the average F1 score calculated per class.  

\begin{table}[htbp]
\renewcommand{\arraystretch}{1.7}
\small
\centering
\begin{tabular}{@{}p{0.4\linewidth}p{0.55\linewidth}@{}}
\toprule
Rhetorical Role Label & Definition \\ \midrule
Fact (FAC) & These are the facts specific to the case based on which the arguments have been made and judgment has been issued. In addition to Fact, we also have the fine-grained label \\
Issues (ISS) &  The issues which have been framed/accepted by the present court for adjudication. \\
Argument Petitioner (ARG-P) &  Arguments which have been put forward by the petitioner/appellant in the case before the present court and by the same party in lower courts (where it may have been petitioner/respondent)\\
Argument Respondent (ARG-R) &  Arguments which have been put forward by the respondent in the case before the present court and by the same party in lower courts (where it may have been petitioner/respondent) \\
Statute (STA) &  The laws referred to in the case. \\
Dissent (DIS) &  Any dissenting opinion expressed by a judge in the present judgment/decision. \\
Precedent Relied Upon (PRE-R) &  The precedents which have been relied upon by the present court for adjudication. These may or may not have been raised by the advocates of the parties and amicus curiae. \\
Precedent Not Relied Upon (PRE-NR) & The precedents which have not been relied upon by the present court for adjudication. These may have been raised by the advocates of the parties and amicus curiae. \\
Precedent Overruled (PRE-O) &  Any precedents (past cases) on the same issue that have been overruled through the current judgment. \\
Ruling By Lower Court (RLC) & Decisions of the lower courts which dealt with the same case. \\
Ratio Of The Decision (ROD) & The principle that has been established by the current judgment/decision which can be used in future cases. Does not include the obiter dicta which is based on observations applicable to the specific case only.  \\
Ruling By Present Court (RPC) & The decision of the court on the issues that have been framed/accepted by the present court for adjudication.  \\
None (NON) & any other matter in the judgment which does not fall in any of the above-mentioned categories.  \\ \bottomrule
\end{tabular}
\caption[RR Label Definitions]{Definitions for different Rhetorical Roles}
\label{tab:rr-definition}
\end{table}







\subsection{Court Judgment Prediction with Explanation (CJPE)} \label{app:cjpe}

\noindent\textbf{Task Motivation and Description:} The task of Court Judgment Prediction with Explanation (CJPE) aims to augment a judge in the judicial decision-making process by predicting the final outcome of the case. Note that the idea behind this task is \textit{not} to replace human judges but to aid them. Furthermore, the task requires the system to explain its decision so that it is interpretable for a human using it. \textbf{Formally, the task of  Court Judgment Prediction with Explanation (CJPE) involves predicting the final judgment (appeal accepted or denied, i.e., the binary outcome of 0 or 1) for a given judgment document (having facts and other details) and providing the explanation for the decision.} The explanations, in this case, are in the form of the crucial sentences appearing in the input text that lead to the decision.

\noindent\textbf{Dataset Details:} For the CJPE task, we use the \texttt{Indian Legal Document Corpus (ILDC)}~\citep{malik2021ildc}. ILDC is a corpus of 34k legal judgment documents (in English) from the Supreme Court of India. Each document is annotated with the ground truth (actual decision given by the judge); further, a small subset of the documents are annotated with explanations by legal experts. This makes it a suitable dataset to consider for a legal understanding benchmark as it covers both judgment as well as relevant explanations annotated by human experts. Table \ref{tab:ildc} provides dataset statistics. Some cases consist of multiple appeals, which can contain corresponding decisions for each appeal. However, since the task has been posed as binary text classification, the final decision is considered as ACCEPT if \textit{at least one} appeal is accepted, otherwise REJECT. The documents are stripped of the final decision given by the Judge with the help of regex-based matching. Table \ref{tab:ildc} provides details of the dataset. 

\noindent Regarding ethical concerns, we follow \citet{malik2021ildc} who took various steps, such as normalizing the dataset concerning named entities to remove any biases in the data (also check the Ethical Considerations section). 


\noindent\textbf{Annotation Details:} The explanation aspect of the CJPE task was annotated with the help of 5 legal experts \cite{malik2021ildc}. The annotators were graduate students and a law professor from a reputed law school. The annotators were not shown the final decision of the case. They were asked to predict the final decision and annotate the sentences (explanations) in the document that led to the final decision. More details about agreement among the annotators are provided in \cite{malik2021ildc}. In a nutshell, the average prediction F1 score of annotators w.r.t. to the ground truth judgment was $94.32\%$. This points towards the challenging nature of the CJPE task; as pointed out earlier, India has a common-law system, and hence, judges could override existing precedents. Disagreements among the annotators were mainly due to differences in the linguistic interpretation of the case and law. For the explanation part, similar trends are reported with the average agreement in terms of the BLEU score to be around $0.4$.  

\noindent\textbf{Evaluation:} The prediction part of the CJPE task is evaluated using standard F1 score metric, and the explanation part is evaluated using BLEU and ROUGE scores. 

\newcommand{\corpusname}{ILDC\xspace}

\newcommand{\taskname}{CJPE\xspace}

\newcommand{\corpuslarge}{\ensuremath{\text{\corpusname}_{\text{multi}}}\xspace}

\newcommand{\corpussmall}{\ensuremath{\text{\corpusname}_{\text{single}}}\xspace}

\newcommand{\corpusexpert}{\ensuremath{\text{\corpusname}_{\text{expert}}}\xspace}


\begin{table}[htbp]
\small
\renewcommand{\arraystretch}{1.1}
\setlength\tabcolsep{15pt}
\hspace{0.2cm}
\centering
\begin{tabular}{p{0.15\columnwidth}p{0.1\columnwidth}p{0.1\columnwidth}p{0.1\columnwidth}}
\toprule
	\multirow{2}{*}{\textbf{\begin{tabular}[c]{@{}c@{}}Corpus\\ (Avg. tokens)\end{tabular}}} &
	\multicolumn{3}{c}{\textbf{\begin{tabular}[c]{@{}c@{}}Number of docs \\ (Accepted Class \%)\end{tabular}}} \\
& \textbf{Train}    & \textbf{Validation} & \textbf{Test}                                                              \\ \midrule
	\textbf{\begin{tabular}[c]{@{}c@{}} ILDC-multi \\ (3231)\end{tabular}} &
	\begin{tabular}[c]{@{}c@{}}32305\\ (41.43\%)\end{tabular} &
		\multirow{3}{*}{\begin{tabular}[c]{@{}c@{}}994\\ (50\%)\end{tabular}} &
		\multirow{3}{*}{\begin{tabular}[c]{@{}c@{}}1517\\ (50.23\%)\end{tabular}} \\
\textbf{\begin{tabular}[c]{@{}c@{}} ILDC-single \\ (3884)\end{tabular}}                     & \begin{tabular}[c]{@{}c@{}}5082\\ (38.08\%)\end{tabular}  &  & \\ \midrule
\textbf{\begin{tabular}[c]{@{}c@{}} ILDC-expert \\ (2894)\end{tabular}}                   & \multicolumn{3}{c}{56 (51.78\%)}                                                                                                                                                                               \\ 

\bottomrule
\end{tabular}
\caption[CJPE Dataset statistics (ILDC)]{Statistics for the CJPE dataset (ILDC) \citep{malik2021ildc}}\label{tab:ildc}
\end{table}


\subsection{Bail Prediction (BAIL)} \label{app:bail}

\noindent\textbf{Task Motivation and Description:} 
A large fraction of the pending cases in India are from the district-level courts, and have to do with bail applications (\url{https://en.wikipedia.org/wiki/Bail})~\citep{kapoor-etal-2022-hldc}. Many of the district courts in India use Hindi as their official language (also refer to the Limitations section). Given the importance of Hindi (the most frequently spoken/written language in India), the task of Bail Prediction for Hindi legal documents is of immense importance, incorporating both language diversity and wider applicability in the Indian legal system. \textbf{Formally, given a legal document (having the facts of the case), the task of Bail Prediction involves predicting if the accused should be granted bail or not (i.e., a binary decision of 0 and 1).} 

\noindent\textbf{Dataset:} 
For the task of BAIL prediction, \citet{kapoor-etal-2022-hldc} created a corpus of 900k Hindi Legal Documents (referred to as HLDC (Hindi Legal Document Copus)). The corpus is created by scrapping publicly available documents on the eCourts website (\url{https://ecourts.gov.in/ecourts_home/}). The documents are scrapped from district courts of the state of Uttar Pradesh (a Hindi-speaking state in northern India). The data is anonymized to take care of biases and ethical aspects; please refer to \cite{kapoor-etal-2022-hldc} for more details. Bail cases in HLDC are pre-processed to remove the final decision (using regex) since we aim to predict this automatically. For the task of Bail prediction we selected only the documents related Bail cases from HLDC, this resulted in 176K documents, having 86 words per document on average. More details about the dataset are discussed in \cite{kapoor-etal-2022-hldc}. For model training and evaluation, we divide the data into train, validation, and test split in the ratio of 70:10:20. 

\noindent\textbf{Evaluation:} The BAIL prediction is a binary task; it is evaluated using the standard macro-F1 score metric. 



\begin{table*}[!t]
\centering
\small
\renewcommand{\arraystretch}{1.3}  
\begin{tabular}{p{0.1\textwidth} p{0.8\textwidth}}
\toprule
{\bf Facts of the case} & 
``On the fateful day at about 9.30 a.m. deceased accompanied by [PERSON1] (PW 4) and [PERSON2] (PW 7) was going from his village Talod to Alote. 
The accused persons were hiding behind bushes on the road near village Gharola. 
They were armed with lathies and farsies.
When the deceased and the aforesaid two persons reached near the Khakhra, the respondents surrounded them and started attacking the deceased with weapons with which they were armed.
His nose was cut. 
PWs. 4 and 7 tried to intervene, but they were also attacked by the accused persons as a result of which they also received injuries. 
The two witness rushed to the police station where PW 4 lodged the FIR (Exhibit P-10). 
The deceased in injured condition was taken to the hospital, and later he succumbed to the injuries. 
Post-mortem was conducted and large number of injuries were found on his body. 
During investigation the alleged weapons of the assailants were seized. After investigation charge sheet was placed.'' \\
\midrule
{\bf IPC S.324} &
{\it Voluntarily causing hurt by dangerous weapons or means}
\\ 
{\bf IPC S.302} &
{\it Punishment for murder}\\
\bottomrule
\end{tabular}%
\caption[LSI Dataset Example (ILSI)]{Example of the LSI task, fact section taken a High Court Document {\bf ``State Of Madhya Pradesh vs. Mansingh And Ors. on 13 August, 2003''}, along with the IPC Sections (324 and 302) that the case cites.} 
\label{app-tab:lsi-example}
\end{table*}

\subsection{Legal Statute Identification (LSI)} \label{app:lsi}

\noindent\textbf{Task Motivation and Description:} 
One of the first steps in the judicial process is finding the applicable statutes/laws based on the facts of the current situation. Manually rummaging through multiple legislation and laws to find out the relevant statutes can be time-consuming, making the LSI task important for reducing the workload, helping improve the efficiency of the judicial system. \textbf{The task of Legal Statute Identification (LSI) is formally defined to automatically identify the relevant statutes given the facts of a case.} An example of the LSI task is presented in the Table~\ref{app-tab:lsi-example}. We utilize the \texttt{ILSI} dataset for this task, which comprises of 100 target statutes from the Indian Penal Code (IPC), the main legislation codifying criminal laws in India.

\noindent\textbf{Dataset:} For LSI, we use the \texttt{Indian Legal Statute Identification (ILSI)} dataset \citep{paul2022lsi}. The dataset consists of fact portions of 65k court case documents (derived from criminal court cases from the Supreme Court of India (SCI) and 6 High Courts of India). The Indian Penal Code (IPC) comprises most criminal statutes and procedures in India; the 100 most frequently occurring statutes in the IPC were chosen as the target statutes. The original ILSI dataset released by \citet{paul2022lsi} contains named entities.
In line with recent works in legal NLP~\citep{malik2021ildc}, we anonymize the dataset by masking entities of types `PERSON' and `ORGANIZATION' to remove any possible bias. Table~\ref{tab:lsi-stats} lists some important statistics about the ILSI dataset. In addition to the facts extracted from case documents and their corresponding statute mappings, \citet{paul2022lsi} also provided the statute descriptions as part of the dataset.

\noindent\textbf{Dataset Preprocessing:}
The LSI task requires the input to be \emph{only} the facts of the case, and thus, an automated RR method~\citep{bhattacharya2019identification} was employed to extract the facts.
Since this method is not foolproof, some sentences containing statute citations may get mislabeled as facts. The version of the dataset released by \citet{paul2022lsi} contains some unmasked statute citations. Thus, we used an existing automated Legal NER method~\citep{kalamkar-etal-2022-named}, which can identify both the act/law names and the statute/section references in the text, to mask all possible statute and act references (statutes from all acts were masked, not just IPC). To prevent model biases, we also masked all entities identified by the Legal NER method.



\noindent\textbf{Evaluation:} LSI is formulated as a multi-label text classification task. 
The facts, a functional segment of the entire case document, are provided as input. 
The expected output is one or more statutes from a list of target statutes relevant to the given fact portion. Standard classification metrics such as macro-averaged precision, recall, and F1 score are used for evaluation. In principle, the LSI task can also be considered a retrieval task instead of a multi-label classification task, i.e., the task is to retrieve from a dynamic set of statutes and provide a bigger pool of relevant candidates to be retrieved for a particular query document having facts. However, as it is an initial phase of establishing the benchmark, we followed the classification setting proposed in previous works~\cite{Wang2018, Wang2019, chalkidis-etal-2019-neural,chalkidis-etal-2021-paragraph}.

\begin{table}[htbp]
\small
\renewcommand{\arraystretch}{1}
\setlength\tabcolsep{1pt}
\centering
\begin{tabular}{p{0.7\columnwidth}c}
\toprule
\textbf{Dataset} & \textbf{ILSI} \\ \midrule
\# Documents & 66,090 \\
\# Labels & 100 \\ 
Train/Dev/Test Split & \begin{tabular}[c]{@{}c@{}c@{}}42,835/\\10,200/\\13,039\end{tabular} \\ 
\midrule
Avg. Document Size (in \#words) & 2406 \\ 
Avg. no. of citations (\#labels per doc) & 3.78 \\
\bottomrule
\end{tabular}
\caption[LSI Dataset statistics (ILSI)]{The table shows the dataset statistics for the ILSI dataset~\citep{paul2022lsi}.}
\label{tab:lsi-stats}
\end{table}


\subsection{Prior Case Retrieval (PCR)} \label{app:pcr}

\noindent\textbf{Task Motivation and Description:} When framing a legal document, legal experts (judges and lawyers) use their expertise to cite previous cases to support their arguments/reasoning. Legal experts have relied on their expertise to cite previous cases; however, with an exponentially growing number of cases, it becomes practically impossible to recall all possible cases. \textbf{Given a query document (without citations), the task of Prior Case Retrieval (PCR) is to retrieve the legal documents from the candidate pool that are relevant (and hence can be cited) in the given query document.} Automating this process directly impacts the justice delivery logistics. Moreover, including this task in the benchmark incorporates the retrieval aspects and understanding of legal similarity (as opposed to semantic similarity), opening research directions for retrieval systems in the legal domain. 

\noindent\textbf{Dataset:} For the task of PCR we use the \texttt{Indian Legal Prior Case Retrieval (IL-PCR)} corpus \citep{joshi-etal-2023-ucreat}. To the best of our knowledge, IL-PCR is the largest publicly available retrieval dataset for the Indian judicial system, making it a suitable candidate to be added to the benchmark. The IL-PCR corpus was created by scraping legal documents (available in the public domain) from the website IndianKanoon (\hyperlink{https://indiankanoon.org/}{https://indiankanoon.org/}). The pool of documents is expanded by scraping documents cited by documents scraped previously. It was done to ensure sufficient citation links from the query to the candidate pool in the final dataset. Names of individuals and organizations were anonymized to the <NAME> and <ORG> tags, respectively, using a NER model \cite{Spacy-io} and a manually compiled gazetteer. This anonymization step is especially pertinent to the PCR task as it removes any biases in the judgment based on entity names. The ground truth labels mark all the candidate's cases relevant to each query case. Statistics for the IL-PCR corpus are shown in Table~\ref{tab:il-pcr-coliee}.

\noindent\textbf{Evaluation:} The PCR task uses micro-averaged F1@K score as the evaluation metric (as done in previous work: \url{https://sites.ualberta.ca/~rabelo/COLIEE2021/}). Prediction models predict a relevance score for each candidate for a given query. Top-K-ranked candidates are considered for prediction (i.e., whether a candidate is cited or not).

\begin{table}[htbp]
\small
\renewcommand{\arraystretch}{1}
\setlength\tabcolsep{10pt}
\hspace{0.2cm}
\centering
\begin{tabular}{lc}
\toprule
\textbf{Dataset}        & \textbf{IL-PCR} \\ \midrule
\# Documents                            &   7070                  \\
Avg. Document Size      &        8093.19             \\
\# query Documents      &         1182            \\
Vocab Size         &    113340                 \\
Total Citation Links       &       8008              \\
Avg. Citation Links per query &      6.775               \\ 
Language &         English              \\
\bottomrule
\end{tabular}
\caption[PCR Dataset Statistics (IL-PCR)]{The table shows the statistics for the IL-PCR dataset~\citep{joshi-etal-2023-ucreat}}
\label{tab:il-pcr-coliee}
\end{table}


\subsection{Summarization (SUMM)} \label{app:summ}


\noindent\textbf{Task Motivation and Description:} Summarization is a standard task in NLP; however, as mentioned in \S \ref{sec:intro}, summarizing legal documents requires legal language understanding and reasoning. \textbf{The task of summarization involves generating a gist (of a legal document) that captures the critical aspects of the case.} Summarization could be extractive (selecting the important sentences) or abstractive (generating the gist). In our setting, summarization is an abstractive generation task. 

\noindent\textbf{Dataset:} For the summarization task, it is necessary to have a large dataset with gold summaries. Consequently, we use the \texttt{In-Abs} dataset \citep{shukla2022summ}, created from judgment documents from the Supreme Court of India. The dataset consists of $7130$ case documents with abstractive summaries (also called “headnotes”). These documents were collected from the website of the Legal Information Institute of India (\url{http://www.liiofindia.org/in/cases/cen/INSC/}), which provides free and non-profit access to databases of Indian law. These documents are accompanied by additional notes called ``headnotes'', which enumerate the important issues and aspects of the case. Legal experts write these headnotes and can be considered abstractive summaries of the entire case document. Headnotes usually occur in the top part of the document, just below the document header (which contains party names, date, bench, etc.), and just above the main judgment. They are also usually preceded by the heading ``HEADNOTE:''. \citet{shukla2022summ} used these cues, and additionally employed regular expression matching to extract the headnotes from the judgment. Table~\ref{tab:summ-stats} provides some statistics of this dataset (more details in \citet{shukla2022summ}).

\begin{table}[htbp]
\small
\renewcommand{\arraystretch}{1}
\setlength\tabcolsep{10pt}
\hspace{0.2cm}
\centering
\begin{tabular}{lc}
\toprule
\textbf{Dataset}        & \textbf{In-Abs} \\ \midrule
\# Documents & 7,130 \\
Type of Summary & Abstractive \\ 
Language & English \\
Train/Test Split & 7,030/100 \\ \midrule
Avg. Document size (in \#words) & 4376.98 \\
Avg. Summary size (in \#words) & 842.52 \\
Avg. Compression Ratio & 0.235 \\

\bottomrule
\end{tabular}
\caption[SUMM Dataset statistics (In-Abs)]{The table shows the statistics of the In-Abs dataset~\citep{shukla2022summ}}
\label{tab:summ-stats}
\end{table}

\noindent\textbf{Evaluation:} Following \citet{shukla2022summ}, we use standard summarization metrics such as ROUGE-1, ROUGE-2, and ROUGE-L F1-scores (computed using \url{https://pypi.org/project/py-rouge/}, with \emph{max\_n} set to 2, parameters \emph{limit\_length} and \emph{length\_limit} not used, and other parameters kept as default), and BertScore~\citep{zhang2019bertscore} (computed using \url{https://pypi.org/project/bert-score/}, version 0.3.4) that calculates the semantic similarity scores using the pre-trained BERT model.


\subsection{Machine Translation (MT)} \label{app:mt}

\noindent\textbf{Task Motivation and Description:} In the Indian legal setting, when a case is transferred (due to re-appeal) from a district court to a High court, the corresponding document (typically in a regional language) needs to be translated to English. 
Additionally, since a large majority of the Indian population is not proficient in English, High Court / Supreme Court documents often need to be translated from English to Indian languages for a better understanding of the involved parties. In both scenarios, such translations, if done by humans, become a primary reason for delay in administering justice. Machine translation (MT) can augment human translators who could post-edit the translated document rather than translating from scratch. As outlined in \S\ref{sec:intro}, legal documents have different lexicons and styles; hence, existing MT systems do not perform well~\cite{mahapatra2023milpac}. Given that many Indian languages are low-resource, MT becomes even more challenging, requiring specialized models for translating legal documents in low-resource Indian languages. \textbf{The task of Legal Machine Translation (L-MT) is to translate text in English to Indian languages and vice-versa.} 

\noindent\textbf{Dataset:} For this task, we use the Multilingual Indian Legal Parallel Corpora (MILPaC)~\citep{mahapatra2023milpac}, which comprises of a total of 17,853 parallel text pairs across English and 9 Indian languages, namely, Bengali (BN), Hindi (HI), Gujarati (GU), Malayalam (ML), Marathi (MR), Telugu (TE), Tamil (TA), Punjabi (PA) and Oriya (OR). MILPaC consists of following 3 datasets:




\noindent{{\bf 1)  MILPaC-IP:} Developed from a set of primers released by a society of law practitioners, this contains a set of approximately 57 question-answer pairs related to Indian Intellectual Property Laws, developed in EN and 9 Indian languages --BN, HI, MR, TA, GU, TE, ML, PA, OR.} The details of the dataset are shown in Table~\ref{tab:mt-stats1}

\noindent{{\bf 2) MILPaC-CCI-FAQ:} is developed from a set of QA booklets released by the Competition Commission of India and contains 184 QA pairs on statutory rules based on competition issues in India. The parallel corpus has been developed for EN and 4 Indian languages --- BN, HI, MA, and TA (see Table~\ref{tab:mt-stats1}).}

\noindent{{\bf 3) MILPaC-Acts:} has been developed from 10 popular Indian Acts (statutory documents outlining laws of the country), for which official translations (from the Indian legislature) were available in English and the 9 Indian languages used in MILPaC-IP. For details, see Table~\ref{tab:mt-stats2}.}

\begin{table}
\small
\centering
\setlength{\tabcolsep}{3pt}
\begin{tabular}{ccccccccccc} 
 \toprule
  & \textbf{EN} & \textbf{BN} & \textbf{HI} & \textbf{MR} & \textbf{TA} & \textbf{TE} & \textbf{ML} & \textbf{PA} & \textbf{OR} & \textbf{GU} \\ 
 \midrule
 \textbf{EN} & $\times$ & 110 & 114 & 114 & 114 & 112 & 114 & 114 & 114 & 114 \\ 
 \textbf{BN} & \textcolor{blue}{\it 365} & $\times$ & 110 & 110 & 110 & 108 & 110 & 110 & 110 & 110 \\
 \textbf{HI} & \textcolor{blue}{\it 365}  & \textcolor{blue}{\it 365} & $\times$ & 114 & 114 & 112 & 114 & 114 & 114 & 114 \\ 
 \textbf{MR} & \textcolor{blue}{\it 365} & \textcolor{blue}{\it 365}  &  \textcolor{blue}{\it 365}& $\times$ & 114 & 112 & 114 & 114 & 114 & 114 \\
 \textbf{TA} & \textcolor{blue}{\it 365} & \textcolor{blue}{\it 365} & \textcolor{blue}{\it 365} & \textcolor{blue}{\it 365} & $\times$ & 112 & 114 & 114 & 114 & 114 \\
 \textbf{TE} &  &  &  &  &  & $\times$ & 112 & 112 & 112 & 112 \\
 \textbf{ML} &  &  &  &  &  &  & $\times$ & 114 & 114 & 114 \\
 \textbf{PA} &  &  &  &  &  &  &  & $\times$ & 114 & 114 \\
 \textbf{OR} &  &  &  &  &  &  &  &  & $\times$ & 114 \\
 \textbf{GU} &  &  &  &  &  &  &  &  &  & $\times$ \\
 \bottomrule
\end{tabular}
\caption[L-MT Dataset Statistics (MiLPaC-IP and MiLPaC-CCI-FAQ)]{Number of parallel text units per language pair in (1)~{\bf MILPaC-IP} - black entries in upper triangular part, and (2)~{\bf MILPaC-CCI-FAQ} - \textcolor{blue}{blue} italicized entries in lower triangular part. For both datasets, text units are QA-pairs, hence not tokenized into sentences (details in text).}
\label{tab:mt-stats1}
\end{table}

\begin{table}[tb]
\small
\centering
\setlength{\tabcolsep}{3pt}
\begin{tabular}{ccccccccccc} 
 \toprule
   & \textbf{EN} & \textbf{BN} & \textbf{HI} & \textbf{MR} & \textbf{TA} & \textbf{TE} & \textbf{ML} & \textbf{PA} & \textbf{OR} & \textbf{GU}\\
 \midrule
 \textbf{EN} & $\times$ & 739  & 706 & 578 & 418 & 319 & 443 & 261 & 256 & 316\\ 
 \textbf{BN} &  & $\times$  & 439 & 439 & $\times$ & 319 & 438 & $\times$ & $\times$ & $\times$\\
 \textbf{HI} &  &  & $\times$ & 578 & $\times$ & 319 & 443 & 262 & 256 & $\times$\\ 
 \textbf{MR} &  &  &  & $\times$ & $\times$ & 319 & 443 & 133 & 128 & $\times$\\
 \textbf{TA} &  &  &  &  & $\times$ & $\times$ & $\times$ & $\times$ & $\times$ & $\times$\\
 \textbf{TE} &  &  &  &  &  & $\times$ & 319 & $\times$  & $\times$ & $\times$\\
 \textbf{ML} &  &  &  &  &  &  &  $\times$ & $\times$  & $\times$ & $\times$\\
 \textbf{PA} &  &  &  &  &  &  &   & $\times$ & 256 & $\times$\\
 \textbf{OR} &  &  &  &  &  &  &  &  & $\times$ & $\times$\\
 \textbf{GU} &  &  &  &  &  &  &  &  &  & $\times$\\
 \bottomrule
\end{tabular}
\caption[L-MT Dataset Statistics (MiLPaC-Acts)]{Number of Parallel Text units per language pair in {\bf MILPaC-Acts}. Text units are tokenized into sentences for this dataset.}
\label{tab:mt-stats2}
\end{table}

\noindent The exact number of pairwise samples are shown in Table~\ref{tab:mt-stats1} (MILPaC-IP and MILPaC-CCI-FAQ) and Table~\ref{tab:mt-stats2}. For more details regarding the creation and curation of the dataset, refer to \citet{mahapatra2023milpac}.

\noindent\textbf{Evaluation:} {Following the evaluation strategies proposed by \citet{mahapatra2023milpac}, we use the standard metrics for machine translation, such as BLEU (Bi-Lingual Evaluation Understudy), GLEU (Google BLEU) and chrF++.
For all metrics, the IndicNLP tokenizer is first used to tokenize the texts in Indian languages. 
For BLEU and chrF++, we use the \textit{SacreBLEU} package (\url{https://pypi.org/project/sacrebleu/}). 
In chrF++ calculation, the default order of character and word n-grams are set to 6 and 2 respectively.
For GLEU, we use the \textit{Huggingface evaluate} library for computation, and consider subsequences containing 1,2,3 and 4 tokens (\url{https://huggingface.co/spaces/evaluate-metric/google_bleu}).}


\subsection{Limitations: Model Explainability}\label{app-sec:limitations}

Model explainability is essential for the legal domain. For some tasks like L-NER, RR, SUMM, and L-MT, based on our discussions with legal practitioners, explainability may not be required as output can easily be observed and interpreted. For tasks like CJPE, BAIL, LSI, and PCR, it is indeed important to know on what basis the model came up with a particular decision. It can be done with the help of various techniques such as occlusion and attention weights (also as done for CJPE); however, for evaluation, such analyses must be verified with legal experts. It requires a significant annotation exercise by legal experts, which is time-consuming and expensive. Nevertheless, in the future, we plan to add more explainability experiments by employing legal experts to annotate explanations for the datasets for tasks like LSI and BAIL. 

\subsection{Ethical Considerations} \label{app-sec:ethics}

\textbf{We do not endorse the use of the benchmark data for non-research (commercial and real-life) applications, and the primary motivation for creating the \benchmarkName\ benchmark is to consolidate all the research happening in parallel for the Indian Legal domain.}  

We took various measures to reduce bias in models trained on legal documents. For relevant tasks (RR, LSI, PCR, CJPE, and BAIL), the documents were anonymized for named entities, judge names and organization names. For example, For anonymizing the ILSI dataset, we used the NER method provided by the paper “Named Entity Recognition in Indian Court Judgments” \cite{kalamkar-etal-2022-named}. We masked entities belonging to categories like PERSON, such as PETITIONER, RESPONDENT, JUDGE, LAWYER, WITNESS, and OTHER\_PERSON. We also masked off entities tagged as PROVISION or STATUTE to remove any mention of statutes from the fact text. According to \citet{kalamkar-etal-2022-named}, their NER model has a macro-F1 of 91.1\%. To further verify the efficacy of the above method on the ILSI dataset, we manually inspected ten randomly selected documents from the test set. We found that over 95\% of entities (belonging to the classes described above) were successfully masked. The model failed in a few cases, e.g., when there was some discrepancy in text formatting, such as no space between a name and a punctuation mark. For CJPE and BAIL tasks, we removed cases (documents) related to sensitive issues like rape and sexual violence, and named entities such as judge names were anonymized. 

Legal data is inherently sensitive and requires careful handling. The automated techniques (since doing it manually is not feasible) used to anonymize the data are not perfect and can sometimes fail. This can possibly have adverse effects. For example, if the names of judges are not anonymized then it can lead to model developing certain spurious correlations (or biases) related to specific type of outcomes associated with a particular judge. Failure to anonymize certain person names (or religion names) can lead to a model developing spurious correlations between certain types of crime and certain religious communities (since certain names are more prevalent in some particular religious communities). Similar things have also been observed in COMPAS system\footnote{\url{https://en.wikipedia.org/wiki/COMPAS_(software)}} for recidivism in the US, where it was biased against certain communities and gender.\footnote{\url{https://www.propublica.org/article/how-we-analyzed-the-compas-recidivism-algorithm}} 
Since Legal-NLP is a relatively new area, to the best of our abilities, we have taken all steps concerning ethical considerations and privacy. Via these tasks, we want to encourage more research in this area so that any hidden factors that could not have been thought of beforehand are also brought to light.



\section{Tasks Models, Experiments and Results} \label{app-sec:results}

In this section, we provide details for all baseline and SOTA models used for each of the tasks.
Apart from these methods, we also conduct inference experiments with LLMs across all of these tasks except PCR, which we discuss in App. \ref{app-sec:llms}. 



\subsection{Legal Named Entity Recognition (L-NER)} \label{app:lner-models}

We perform NER based on token representations generated by BERT-based models. Since each document in the dataset does not come pre-segmented into sentences or paragraphs, we need to chunk documents before passing them to BERT, as case documents easily exceed the token limits of BERT.
However, unlike other tasks like text classification, we need to devise a chunking strategy to avoid splitting true NEs into different chunks. 
For this, we choose to chunk at the last stopword (based on NLTK's list of English stopwords), which satisfies the chunk size limit.
The assumption is that these stopwords are not expected to be part of entity names.

We experiment with five different BERT encoders: (i)~\texttt{bert-base-uncased}~\citep{devlin-etal-2019-bert}, (ii)~LegalBERT~\citep{chalkidis2020legalbert}, (iii)~CaseLawBERT~\citep{zheng2021caselaw}, (iv)~InLegalBERT~\citep{paul2023pretrained} and (v)~InCaseLawBERT~\citep{paul2023pretrained}.
We applied a Conditional Random Field (CRF) on top of the BERT encoder due to the efficacy of CRFs in sequence labeling tasks.

\noindent\textbf{Hyper-parameter Settings:}
We set the chunk limit to 512 tokens to maximize the input capability of BERT. We trained on a single Nvidia RTX A6000 (48 GB). 
We used a batch size of 40 during training and 24 during testing.
The models were trained for a maximum of 20 epochs with early stopping.
We used different learning rates for the different layers, viz., 3e-5 for the BERT layers and 1e-3 for the fully connected and CRF layers.
We have used the PyTorch implementation of CRF provided in \url{https://pypi.org/project/pytorch-crf/}.

\noindent\textbf{Model Result and Analysis:} Since the dataset is small, we divide the 105 documents into three folds (by trying to maintain the class label frequency distribution across folds as much as possible). We perform 3-fold cross-validation and report the mean across folds.
In addition to the strict scores, we also consider another type of scoring, called \textit{ent-type} score~\citep{segura-bedmar-etal-2013-semeval}. This scheme considers a match correct if the predicted label type is the same as that of the ground truth, \emph{even if} the predicted span is not correct. Naturally, this scheme is more lenient than the strict mechanism. We report both strict and ent-type scores for all models in Table~\ref{tab:ner-results}.

\begin{table}
    \centering
    \footnotesize
    \renewcommand{\arraystretch}{1.3}
    \setlength\tabcolsep{2.5pt}
    \hspace{0.2cm}
    \begin{tabular}{l| c  c  c |c  c  c}
       
        \hline
        \multirow{2}{*}{\textbf{Method}} & \multicolumn{3}{c|}{\textbf{Strict}}& \multicolumn{3}{c}{\textbf{Ent type}}\\
        & \textbf{mP} & \textbf{mR} & \textbf{mF1} & \textbf{mP} & \textbf{mR} & \textbf{mF1}\\ \hline
        BERT& 38.95& 41.12& 39.59& 47.70& 49.99& 48.23\\ 
        LegalBERT& 43.98& 48.06& 45.58& 53.19& 58.33&55.21\\
        CaseLawBERT& 42.68& 43.68& 42.45& 52.40& 53.48&52.00\\
        InLegalBERT& \textbf{47.83}& \textbf{50.33}& \textbf{48.58}& \textbf{57.45}& \textbf{60.40}&\textbf{58.30}\\
        InCaseLawBERT& 45.59& 44.59& 44.17& 56.38& 54.89&54.41\\\hline
    \end{tabular}
    \caption[L-NER Task Results]{Performance of BERT-based models over the L-NER dataset. All values are macro-averaged and in terms of percentage.}
    \label{tab:ner-results}
\end{table}


In terms of F1 scores, all the models perform relatively poorly. 
The L-NER dataset contains entire case documents, and evaluation is done over \textit{every occurence} of every named entity. 
This means that models cannot always rely on the local context to infer the nature of an entity, and all these models are incapable of long range context modeling since the inputs are chunked before feeding to them. This could be a possible reason for the low results.
For every model, the ent-type scores are around 20\% higher than the strict scores, suggesting that these models also struggle to identify the NE boundaries correctly on quite a few occasions.
Comparing among the models, we observe increasing performance with greater degree of domain familiarization. BERT performs the poorest, followed by LegalBERT and CaseLawBERT (which have been pre-trained on legal data from other countries).
Counterparts for these models pre-trained on Indian legal text, viz., InLegalBERT and InCaseLawBERT, further outperform them.

\begin{table*}[htbp]
\small
\renewcommand{\arraystretch}{1}
\setlength\tabcolsep{3.5pt}
\hspace{0.2cm}
\centering
\begin{tabular}{lccccccccccccc}
\toprule
\textbf{Label} & \textbf{APP} & \textbf{RESP} & \textbf{JUD} & \textbf{AC} & \textbf{RC} & \textbf{CRT} & \textbf{AUTH} & \textbf{WIT} & \textbf{STAT} & \textbf{PREC} & \textbf{DATE} & \textbf{CN} & \textbf{Macro} \\ \midrule
Strict F1 & 22.72& 11.70& 57.33& 61.27& 53.32& 69.16& 44.37& 29.32& 63.45& 36.99& 81.52& 51.76& 48.58\\
Ent-type F1 & 34.14& 18.01& 71.30& 67.18& 58.80& 76.09& 50.13& 34.21& 72.43& 64.49& 85.06& 67.73& 58.30\\
\bottomrule
\end{tabular}
\caption[L-NER Label-wise Results]{Performance of the best model (InLegalBERT) across all labels of the L-NER dataset. All values are in terms of macro-F1(percentage).}
\label{tab:ner-label-analysis}
\end{table*}

\noindent\textbf{Label Analysis:}
To further analyze the performance across different labels, we calculate the strict and ent-type F1 scores of every label of the best-performing model, InLegalBERT. 

Labels like WITNESS, A.COUNSEL, and R.COUNSEL are straightforward to identify, possibly due to the presence of linguistic cues like ``P.W.'' (abbreviated for ``Prosecution Witness'') and ``learned counsel for the appellant/respondent'' close to the entity mentions. 
Labels like COURT, AUTHORITY, and DATE are slightly more challenging to identify due to the large degree of variations possible in the way these entities are mentioned, e.g., ``Delhi High Court'' vs. ``High Court of Judicature at New Delhi'', or ``14.06.2023'' vs. ``14/6/23'' vs. ``14th June 2023''. We also observe very little difference in these classes' strict and ent-type scores.

Labels like APPELLANT, RESPONDENT, and JUDGE are more challenging to identify. There is an apparent confusion between APPELLANT and RESPONDENT roles since the entities belonging to these classes usually occur in the same context and play the same role in the court case (just opposing sides). However, the performance of the JUDGE class is lower, although JUDGE type entities are usually enclosed by prefixes such as ``Honourable Justice'' or suffixes such as ``J.''. The considerable difference in the strict and ent-type scores for the JUDGE class indicates that the model fails to detect the spans properly rather than the class.

Finally, for labels like STAT, PREC, and CASE NO., the spans can be challenging to identify even for human readers since these entities are usually long, occur in multiple forms, and can have extended suffixes. For example, STAT can either be in the full form, such as ``Indian Penal Code, 1860'' or its abbreviated version ``I.P.C.'', while PREC entities can sometimes contain the case number of the particular precedent as a suffix. The considerable differences in strict and ent-type scores of these entities also point to this possibility. 

\subsection{Rhetorical Role Prediction (RR)} \label{app:rr-models}

For the task of RR Prediction, we experiment with different approaches, such as passing each sentence individually to BERT, LegalBERT and InLegalBERT or applying hierarchical approaches to model the entire document together, such as BiLSTM-CRF with sent2vec~\citep{gupta2019sent2vec} or BERT embeddings.
\citet{malik-rr-2021} suggest an auxiliary task, Label Shift Prediction (LSP), which aims to predict, for sentence $i$ in a document, whether the label changed from sentence $i-1$ to $i$.
This is based on the intuition that RRs tend to maintain some inertia when going from one sentence to another, and changes in RR labels are not abrupt but smooth.
BERT-SC is obtained by fine-tuning BERT for the LSP task \textit{only} over the train set of the RR dataset.
Finally, the Multi-task Learning (MTL) approach incorporates both RR (main task) and LSP (auxiliary task) prediction. For more details about LSP and MTL, check \citet{malik-rr-2021}.

\begin{table}[htbp]
    \centering
    \small
    \begin{tabular}{lccc}
    \toprule \textbf{Model} & \textbf{IT} & \textbf{CL} & \textbf{IT+CL} \\
    \midrule BERT & 0.56 & 0.52 & 0.58 \\
     LegalBERT & 0.55 & 0.53 & 0.56\\
     InLegalBERT & 0.64 & 0.52 & 0.58 \\
     BiLSTM-CRF (sent2vec) & 0.59 & 0.61 & 0.60 \\
     BiLSTM-CRF (BERT emb) & 0.63 & 0.63 & 0.63 \\
     LSP (BERT-SC) & 0.65 & 0.68 & 0.67 \\
     MTL (BERT-SC) & \textbf{0.70} & \textbf{0.69} & \textbf{0.70} \\
    \bottomrule
    \end{tabular}
    \caption[RR Task Results]{Performance of different models on the RR dataset. All values are in terms of Macro-F1.}
    \label{tab:rr-task-results}
\end{table}


\noindent\textbf{Results and Analysis:} Table~\ref{tab:rr-task-results} compares the performances of different models for the RR task. It is evident that RR prediction is a challenging task; standard transformer-based models like BERT, LegalBERT and InLegalBERT applied on individual sentences do not perform well. 
Posing the task as a sequence labeling problem, the hierarchical models employing BiLSTM-CRF show improvements.
LSP plays a significant role in improving performance, which is seen in the performance of LSP (BERT-SC) over models that do not employ LSP.
Harnessing the power of learning both RR and LSP prediction in an end-to-end setup, the MTL model performs the best.
However, this is still quite far from human annotations, pointing towards significant scope for improvement. 

\begin{table*}[htbp]
    \centering
    \small
    \begin{tabular}{lcccc}
\hline \textbf{Model} & \begin{tabular}{l} 
\textbf{Macro} \\
\textbf{Precision} \\
$(\%)$
\end{tabular} & \begin{tabular}{l} 
\textbf{Macro} \\
\textbf{Recall} \\
$(\%)$
\end{tabular} & \begin{tabular}{l} 
\textbf{Macro} \\
\textbf{F1} \\
$(\%)$
\end{tabular} & \begin{tabular}{l} 
\textbf{Accuracy} \\
$(\%)$
\end{tabular} \\
\hline BERT & 69.33 & 67.31 & 68.31 & 67.24 \\
RoBERTa & 72.25 & 71.31 & 71.77 & 71.26 \\
XLNet & 72.09 & 70.07 & 71.07 & 70.01 \\
InLegalBERT & 74.13 & 73.86 & 73.76 & 72.87 \\
\hline BERT + BiGRU & 70.98 & 70.42 & 70.69 & 70.38 \\
RoBERTa + BiGRU & 75.13 & 74.30 & 74.71 & 74.33 \\
XLNet + BiGRU & 77.80 & 77.78 & 77.79 & 77.78 \\
\hline BERT + BiGRU-attn & 71.31 & 70.98 & 71.14 & 71.26 \\
RoBERTa + BiGRU-attn & 75.89 & 74.88 & 75.38 & 74.91 \\
XLNet + BiGRU-attn & 77.32 & 76.82 & 77.07 & 77.01 \\
LegalBERT + BiLSTM-attn & 77.73 & 77.02 & 76.90 & 77.06 \\
InLegalBERT + BiLSTM-attn & 82.15 & 81.45 & 81.31 & 81.41 \\
\hline
\end{tabular}
\caption[CJPE Prediction Results]{Performance of different models on the ILDC-multi dataset. All values are macro-averaged and in terms of percentage.}
    \label{tab:cjpe-results}
\end{table*}

\subsection{Court Judgment Prediction with Explanation (CJPE)} \label{app:cjpe-models}

We use the \texttt{ILDC-multi} split for judgment prediction and \texttt{ILDC-expert} for explanations, out of the ILDC dataset developed by~\cite{malik2021ildc}.
Different transformer-based models (BERT, RoBERTa and XLNet, InLegalBERT) have been tried for the CJPE task.
Since these models cannot accommodate large documents, one approach is to make the prediction based on a chunk of 512 tokens.
The last 512 tokens are chosen since these parts of the text are likely to contain more information for guiding the final decision~\cite{malik2021ildc}.
In other settings, a hierarchical approach is adopted by chunking the entire document into chunks of 512 tokens, passing these to the transformer, and collecting the [CLS] embeddings to be fed to a high-level encoder, such as BiGRU or BiGRU coupled with attention.

For the explanation part, an occlusion method is used by \citet{malik2021ildc}. The primary idea behind this is to mask a chunk of text and then see the change in prediction probability. The prediction probability change indicates the salience of that particular chunk for making the prediction. The more the change in probability, the more salient the chunk.


\noindent\textbf{Results and Analysis:} From Table~\ref{tab:cjpe-results}, it is evident that the hierarchical models perform better than their counterparts that take just the last 512 tokens (and thus suffer from loss of information). While adding the attn. The layer to the BiGRU module seems to help BERT and RoBERTa slightly, but the same is not true for XLNet. 
However, BERT-based models developed for the Indian legal domain, such as InLegalBERT~\citep{paul2023pretrained}, outperform the open-domain encoders and achieve state-of-the-art performance in terms of macro F1.

The occlusion approach for extracting explanations can give positive or negative scores to each chunk; we choose the chunks that obtain positive scores. The text from these chunks is concatenated and compared with the expert-annotated chunks (5 different annotations for 5 experts).
We only consider all sentences ranked 1 to 10 by the experts as gold-standard explanations (note that many sentences are not ranked).
As described by ~\citet{malik2021ildc}, the occlusion scores are calculated at the chunk level from the hierarchical model and at the sentence level (for a particular chunk) from the flat model. Thereon, chunks with positive score are chosen, and among them, the top 50\% sentences in terms of occlusion score are chosen for evaluation w.r.t. gold-standard.
The best model, InLegalBERT + BiLSTM-Attn, gives 0.561 Rouge-L score and 0.325 BLEU score averaged across all experts.
This demonstrates that explainability is still a big challenge, and the model's understanding of important sentences is quite far off from that of the experts.

\begin{table}[htbp]
    \centering
    \small
    \setlength\tabcolsep{4pt}
    \begin{tabular}{lll}
    \toprule \textbf{Model} & \textbf{Accuracy} & \textbf{F1} \\\midrule
    IndicBert-First 512 & 0.73 & 0.71 \\
    IndicBert-Last 512 & 0.78 & 0.76 \\
    TF-IDF+IndicBert & $\mathbf{0.82}$ & $\mathbf{0.81}$ \\
    TextRank+IndicBert & 0.82 & 0.81 \\
    Salience Pred.+IndicBert & 0.80 & 0.78 \\
    Multi-Task & 0.80 & 0.78 \\
    \bottomrule
    \end{tabular}

    \caption[BAIL Task Results]{Performance of different models over the HLDC dataset. F1 values are macro-averaged, and all values are in terms of percentage.}
    \label{tab:bail-results}
\end{table}

\subsection{Bail Prediction (BAIL)} \label{app:bail-models}

We use the \texttt{HLDC-all-districts}~\citep{kapoor-etal-2022-hldc} split for all our experiments.
For BAIL, we used the multi-lingual IndicBERT~\citep{kakwani2020indicnlpsuite} to encode the facts and predict. Since the facts can be long, some unsupervised summarization-based approaches (such as TF-IDF ranking and TextRank) have been tried to shorten the inputs and remove noise.
We also experiment with the salience prediction approach demonstrated by \citet{kapoor-etal-2022-hldc} that aims to predict the important sentences via supervised learning of salience scores (the gold standard scores are decided by comparing each fact sentence with the final case summary written by the judge).
Finally, we also an MTL approach by combining BAIL and salience prediction tasks is also carried out.

\noindent\textbf{Results and Analysis:} The results are reported in Table~\ref{tab:bail-results}. As we observe, summarization of the input facts is a better approach than just taking the first or last 512 tokens for passing to IndicBERT.
Surprisingly, TF-IDF shows the best performance with 81\% macro-F1, even outperforming supervised salience prediction and MTL approaches.
This could possibly be because of the large variation in the nature and dialect of text across the entire dataset.

\subsection{Legal Statute Identification (LSI)} \label{app:lsi-models}
We chose some models from the BERT family -- LegalBERT~\citep{chalkidis2020legalbert} and InLegalBERT~\citep{paul2023pretrained} as baselines for this task. 
Since fact descriptions (input for LSI) can be long, they may not fit within the maximum 512-token limit for BERT encoders, necessitating a hierarchical model. Examples from the ILSI dataset are pre-segmented into sentences. We pass each sentence individually through the BERT encoder and gather the [CLS] embeddings for each document. 
The sequence of [CLS]-embeddings are passed through an upper Bi-LSTM layer coupled with attention, yielding a single representation for the entire fact portion. It then passes through a fully connected layer with sigmoid activation to obtain label probabilities.
Labels with a probability score $> 0.5$ are considered relevant.
Apart from these two models, we also experiment with LeSICiN~\citep{paul2022lsi}, a graph-based deep neural model that also utilizes sent2vec~\citep{gupta2019sent2vec} embeddings pre-trained on Indian legal data.


\begin{table}[htbp]
    \centering
    \small
    \renewcommand{\arraystretch}{1.3}
    \setlength\tabcolsep{4pt}
    \begin{tabular}{lccc}
        \hline
        \textbf{Encoder Module} & \textbf{mP} & \textbf{mR} & \textbf{mF1} \\\hline
        LegalBERT + LSTM-Attn & 53.79& 15.72& 21.74\\
        InLegalBERT + LSTM-Attn & \textbf{58.75} & 19.29 & 26.23 \\
        LeSICiN & 24.34 & \textbf{36.58} & \textbf{28.08} \\ \hline
    \end{tabular}
    \caption[Performance for the LSI Task]{Performance over the \texttt{ILSI} dataset for LSI. All reported values are macro-averaged and in terms of percentage.}
    \label{tab:lsi-results}
\end{table}

\noindent\textbf{Results:}
The results are reported in Table~\ref{tab:lsi-results}. 
All models perform poorly, indicating the challenging nature of the ILSI dataset.
Among the BERT-based methods, InLegalBERT outperforms LegalBERT since the former has been trained on Indian legal documents and is likely to have more inherent domain knowledge.
While the BERT-based methods utilize strong contextual representations to identify patterns in the fact text that highly correlate with certain labels (high precision), the low recall suggests that the model is not able to pick up more latent patterns.
On the other hand, LeSICiN shows a comparatively better recall since it compares the fact text with the text of the statutes via a graph neural network but has poor precision. Overall, LeSICiN still manages to outperform the BERT-based methods.



\begin{table*}[htbp]
\setlength\tabcolsep{7pt}
\centering
\small
\begin{tabular}{cccccc}
\toprule 
\multicolumn{2}{c}{\textbf{Model}}                                                       &
\textbf{K} &
\textbf{Precision} &
\textbf{Recall} &
\textbf{F1}
\\ \midrule
\multirow{2}{*}{Word Level}                & BM25                                             & 5 &   17.11    & 11.64 & 13.85        \\
                                           & BM25 (Bigram)                                   & 7 &  29.30  & 27.91 & 28.59           \\
                                           \midrule
\multirow{6}{*}{\begin{tabular}{@{}c@{}}Segmented-Doc  \\ Transformer \\(full document) \end{tabular}}     
& BERT               & 6 &   10.28   & 8.40 & 9.24             \\
                                          & BERT (finetuned)    & 6 &    8.79    &        7.18 & 7.90       \\
                                          & DistilBERT                       & 7 &         17.02         & 16.21  & 16.61             \\
                                          & DistilBERT (finetuned)            & 5 &       9.70         &    6.60  & 7.86         \\
 & LegalBERT& 6& 7.87& 9.65&8.67\\
                                          & InCaseLawBERT     & 11 &        3.02             & 4.52  & 3.62             \\
                                          & InLegalBERT     & 12 &          6.10            & 9.96 & 7.56                \\
                                          \midrule
                                          
                                          
\multirow{4}{*}{Atomic Events} & Jaccard Similarity & 7 & 35.12 & 33.28 & 34.17  \\ & BM25                                           & 7 &    37.69    & 35.90 & 36.77       \\
                                           & BM25 (Bigram)                         & 6 &     35.39    & 28.89 & 31.81        \\
                                           & BM25 (Trigram)                                & 6 &     30.71   &  25.07 & 27.61       \\
                                           
                                           \midrule

\multirow{5}{*}{Events Filtered Docs}      & BM25                                   & 5 &    24.26   & 16.50   & 19.64       \\
                                           & BM25 (Bigram)                                  & 6 &    33.69   & 27.50  & 30.28           \\
                                           & BM25 (Trigram)                       & 6 &    \textbf{41.35}   & 33.76 & 37.17        \\
                                           & BM25 (Quad-gram)                               & 7 &  40.12   & \textbf{38.22}  & \textbf{39.15}  \\
                                           & BM25 (Penta-gram)                              & 7 &     39.57     & 37.70 & 38.61       \\
\bottomrule
\end{tabular}

\caption[PCR Task Results]{Performance of different models on the IL-PCR dataset. The table shows the K values, and Precision, Recall, and F1 scores (in terms of percentage) for each model.}
\label{tab:pcr-results}
\end{table*}

\subsection{Prior Case Retrieval} \label{app:pcr-models}

For the PCR task, a classical IR baseline BM-25, apart from some transformer-based approaches, is chosen. We follow the baselines proposed in \citep{joshi-etal-2023-ucreat} and perform all the experiments, including the ones where a document is converted to a set of events.

\noindent\textbf{Results and Analysis:}
The results are shown in Table~\ref{tab:pcr-results}.
BM-25 seems to be a strong baseline, and BERT-based models fail to outperform this. 
In fact, the scores of transformer-based approaches are surprisingly low (less than 10\% F1).
Instead, the event-filtered doc approach works the best.
Comparing the two event-based approaches, working directly with the atomic events works better for BM25 approaches with unigrams and bigrams, but for trigram onwards, the event-filtered doc approach outperforms this.

We have observed that the event-based models perform the best but still have a micro F1 score of 39.15, which is relatively low. Given the low scores, there is massive scope for developing better models for PCR. 
 


\subsection{Summarization (SUMM)} \label{app:summ-models}

Although the \texttt{IN-Abs} dataset is meant for abstractive summarization, we can apply both extractive and abstractive methods \citep{shukla2022summ}. 

\noindent (i)~\textbf{Extractive methods:} We try out approaches like CaseSummarizer~\citep{polsley2016casesummarizer} (legal-specific, unsupervised), DSDR~\citep{he2012dsdr} (open domain, unsupervised), Gist~\citep{liu2019gist} (legal-specific, supervised) and SummaRuNNer~\citep{nallapati2017summarunner} (open domain, supervised). To adapt the abstractive gold-standard summaries for these extractive methods, we use the technique suggested by ~\citet{narayan2018ranking}.

\noindent (ii)~\textbf{Fine-tuned Abstractive methods:} We try out text generation models both from the open-domain like BART~\cite{lewis2019bart}, and legal domain like Legal-Pegasus \cite{legal-pegasus} and Legal-LED \cite{legal-led}. 
While Legal-LED can accommodate a large number of documents (16,384 token limit), the same is not true for the other models. 
To overcome this problem, we chunk the document into equal-sized chunks (each chunk size is lesser than the model length limit) and pass each chunk through the model. The summaries for each chunk are concatenated to form the final summary. To convert the overall document summary (gold standard) into chunk-wise summaries, we follow the approach given by \citet{gidiotis2020summ}. All the models were fine-tuned on the train part of the IN-Abs dataset.

\noindent\textbf{Model Result and Analysis}
The results of all approaches are reported in Table~\ref{tab:summ-results}. SummaRuNNer performs the best among the extractive approaches across three of the four metrics considered (Rouge-1 \& 2, and BERTScore). The abstractive approaches show a general improvement over the extractive ones, possibly due to the gold-standard summaries also being abstractive. Despite being open-domain and requiring chunking, the BART model still comes close to or outperforms Legal-LED across different legal domain-specific metrics and can accommodate very long documents. Legal Pegasus beats BART in terms of R-2 and R-L but falls short in terms of R-1. Legal-LED outperforms every other model in terms of BERTScore.

\begin{table}[htbp]
\small
\centering
\renewcommand{\arraystretch}{1.3}
\setlength\tabcolsep{3pt}
\begin{tabular}{l c c c c}
\toprule
\multirow{2}{*}{\textbf{Algorithm}} & \multicolumn{3}{c}{\textbf{ROUGE Scores}} & \multirow{2}{*}{\textbf{BERTScore}} \\ 
& R-1 & R-2 & R-L & \\ \midrule

\multicolumn{5}{c}{\textit{Extractive Methods (U: Unsupervised, S: Supervised)}} \\ \midrule

DSDR (U) & 0.485 & 0.222 & 0.270 & 0.848 \\ 
CaseSummarizer (U) & 0.454 & 0.229 & 0.279 & 0.843 \\ 
SummaRunner (S) & 0.493 & \textbf{0.255} & 0.274 & 0.849 \\ 
Gist (S) & 0.471 & 0.238 & 0.308 & 0.842 \\ \midrule

\multicolumn{5}{c}{\textit{Abstractive Methods}} \\ \midrule
BART & \textbf{0.495} & 0.249 & 0.330 & 0.851 \\ 
Legal-Pegasus & 0.488 & 0.252 & \textbf{0.341} & 0.851 \\ 
Legal-LED & 0.471 & 0.235 & 0.332 & \textbf{0.856} \\ \midrule


\end{tabular}
\caption[SUMM Task Results]{Document-wide ROUGE-L and BERTScores (Fscore) on the IN-Abs dataset, averaged over the $100$ test documents.}
\label{tab:summ-results}
\end{table}

\noindent \begin{table*}[!htb]
\centering
\small
\begin{tabular}{cc|ccc|ccc|ccc}
\toprule
\multirow{2}{*}{EN $\rightarrow$ IN} & Model & \multicolumn{3}{c|}{{\bf MILPaC-IP}} & \multicolumn{3}{c|}{{\bf MILPaC-Acts}} & \multicolumn{3}{c}{{\bf MILPaC-CCI-FAQ}} \\
& & BLEU & GLEU & chrF++ & BLEU & GLEU & chrF++ & BLEU & GLEU & chrF++\\
\midrule
\multirow{3}{*}{EN $\rightarrow$ BN} & GOOG & 27.7 & 30.7 & 56.8 & 12.0 & 17.0 & 40.7 & \textbf{52.0} & \textbf{53.6} & \textbf{74.8}\\
& MSFT & \textbf{31.0} & \textbf{33.8} & \textbf{59.4} & 18.4 & \textbf{23.1} & \textbf{45.6} & 36.5 & 40.4 & 66.2 \\
& IndicTrans & 24.7 & 27.3 & 51.7 & \textbf{18.6} & 21.8 & 45.5 & 20.9 & 25.6 & 50.2 \\
\midrule
\multirow{3}{*}{EN $\rightarrow$ HI} & GOOG & 36.6 & 35.3 & 53.8 & 21.2 & 26.7 & 47.1 & 46.0 & 48.4 & 67.3\\
& MSFT & \textbf{38.5} & \textbf{37.0} & \textbf{54.9} & \textbf{46.4} & \textbf{48.9} & \textbf{67.3} & 45.5 & 48.2 & \textbf{67.5} \\
& IndicTrans & 27.0 & 28.1 & 45.1 & 45.7 & 48.2 & 66.6 & \textbf{49.1} & \textbf{49.8} & 67.1 \\
\midrule
\multirow{3}{*}{EN $\rightarrow$ TA} & GOOG & \textbf{39.3} & \textbf{41.8} & \textbf{69.4} & 8.1 & 13.7 & 37.0 & \textbf{41.4} & \textbf{44.0} & \textbf{70.7}\\
& MSFT & 35.3 & 38.7 & 68.8 & \textbf{12.1} & \textbf{17.6} & \textbf{46.3} & 29.5 & 33.7 & 64.9 \\
& IndicTrans & 21.4 & 25.5 & 51.9 & 11.1 & 16.7 & 43.7 & 22.9 & 26.8 & 56.1 \\
\midrule
\multirow{3}{*}{EN $\rightarrow$ MR} & GOOG & \textbf{23.0} & \textbf{25.6} & \textbf{51.6} & 8.6 & 14.6 & 37.5 & \textbf{51.3} & \textbf{53.0} & \textbf{74.8}\\
& MSFT & 19.4 & 22.8 & 49.6 & \textbf{13.9} & \textbf{19.6} & \textbf{45.0} & 34.1 & 38.3 & 65.8 \\
& IndicTrans & 16.0 & 19.6 & 44.0 & 12.9 & 18.5 & 42.1 & 28.2 & 32.0 & 56.7 \\
\midrule
\multirow{3}{*}{EN $\rightarrow$ TE} & GOOG & \textbf{22.4} & \textbf{23.2} & \textbf{48.9} & 6.6 & 11.4 & 28.8 &  - &  - & - \\
& MSFT & 15.8 & 18.3 & 44.8 & \textbf{12.0} & \textbf{16.9} & 39.4 &    - &  - & - \\
& IndicTrans & 15.5 & 17.6 & 40.6 & 11.9 & 16.8 & \textbf{40.4} & - &  - & - \\
\midrule
\multirow{3}{*}{EN $\rightarrow$ ML} & GOOG & 22.3 & 27.7 & 57.5 & 7.3 & 12.4 & 32.2 &    - &  - & - \\
& MSFT & \textbf{34.2} & \textbf{37.7} & \textbf{66.5} & 10.8 & 17.0 & 46.2 &   - &  - & - \\
& IndicTrans & 19.8 & 24.5 & 48.9 & \textbf{16.6} & \textbf{21.2} & \textbf{50.3} & - &  - & - \\
\midrule
\multirow{3}{*}{EN $\rightarrow$ PA} & GOOG & 17.8 & 20.8 & 41.3 & 8.9 & 14.1 & 28.6 &    - &  - & - \textbf{}\\
& MSFT & \textbf{30.2} & \textbf{30.5} & \textbf{51.3} & \textbf{40.1} & \textbf{42.4} & \textbf{62.5} & - &  - & - \\
& IndicTrans & 28.1 & 28.8 & 47.6 & 24.0 & 28.8 & 48.8 & - &  - & - \\
\midrule
\multirow{3}{*}{EN $\rightarrow$ OR} & GOOG & 2.4 & 6.5 & 29.0 & 4.1 & 8.2 & 26.3 &   - &  - & - \\
& MSFT & 5.5 & 9.0 & 33.7 & 7.6 & 13.3 & 37.3 &   - &  - & - \\
& IndicTrans & 4.9 & 8.6 & 30.5 & 8.9 & 15.0 & \textbf{40.4}  & - &  - & - \\
\midrule
\multirow{3}{*}{EN $\rightarrow$ GU} & GOOG & 43.6 & 46.0 & 67.8 & 14.3 & 19.5 & 42.1 & - &  - & - \\
& MSFT & \textbf{47.3} & \textbf{49.2} & \textbf{70.6} & 21.7 & 26.1 & \textbf{51.9} & - &  - & - \\
& IndicTrans & 31.3 & 34.9 & 56.3 & \textbf{22.9} & \textbf{27.0} & 50.9 & - &  - & - \\
\midrule
\multirow{3}{*}{Average} & GOOG & 26.1 & 28.6 & 47.6 & 10.1 & 15.3 & 35.6 & \textbf{47.7} & \textbf{49.8} & \textbf{71.9}\\
& MSFT & \textbf{28.6} & \textbf{30.8} & \textbf{55.5} & 20.3 & 25.0 & 49.1 & 36.4 & 40.2 & 66.1 \\
& IndicTrans & 24.4 & 27.8 & 52.5 & \textbf{21.7} & \textbf{26.8} & \textbf{53.3} & 30.3 & 33.6 & 57.5 \\
\bottomrule
\end{tabular}
\caption[L-MT Task Dataset-wise Results]{Corpus-level BLEU, GLEU, and chrF++ scores for all MT systems, over three datasets. All values are averaged over all text pairs in a particular dataset. For each dataset and each English-Indian language pair, the best value of each metric is boldfaced.}
\label{tab:mt-results}
\end{table*}

\subsection{Legal Machine Translation (L-MT)} \label{app:mt-models}

{For this task, we employed a host of systems, including Commercial systems such as Google Cloud Translation - Advanced Edition (v3) system\footnote{\url{https://cloud.google.com/translate/docs/samples/translate-v3-translate-text}} (\textbf{GOOG}) and the Translation API offered by Microsoft Azure Cognitive Services (v3)\footnote{\url{https://azure.microsoft.com/en-us/products/cognitive-services/translator}} (\textbf{MSFT}).
We also used open-source models such as IndicTrans, which is a transformer-4x based multilingual NMT model\footnote{\url{https://github.com/AI4Bharat/indicTrans}} trained over the \textit{Samanantar} dataset for translation among Indian languages~\citep{ramesh2022samanantar}.}

\noindent 
\begin{table}[!t]
\centering
\small
\setlength\tabcolsep{3pt}
\begin{tabular}{c|c|c|c}
\toprule
Model & BLEU & GLEU & chrF++ \\ \midrule
GOOG & 28.0 & 31.2 & 51.7 \\
MSFT & \textbf{28.4} & \textbf{32} & \textbf{56.9} \\
IndicTrans & 25.5 & 29.4 & 54.4 \\ \bottomrule
\end{tabular}
\caption[L-MT Task results]{Corpus-level BLEU, GLEU, and chrF++ scores for all MT systems. All values are averaged over all text pairs, across all languages, and across 3 datasets.}
\label{tab:mt-allresults}
\end{table}

\noindent\textbf{Model Result and Analysis}
{The performances of all the MT systems across the 3 datasets are presented in Table~\ref{tab:mt-results}.
We find that no single model performs the best in all scenarios.
MSFT, GOOG, and IndicTrans are the 3 best models that generally perform the best in most scenarios.
The scores for {\bf MILPaC-Acts} are consistently lower than those for other datasets. This is expected since {\bf MILPaC-Acts} has very formal legal language, which is challenging for all MT systems. Interestingly, though MSFT and GOOG perform the best over most datasets, IndicTrans performs better over {\bf MILPaC-Acts} for several Indian languages (e.g., Malayalam \& Gujarati). 
The superior performance of IndicTrans over {\bf MILPaC-Acts} may stem partly from the fact that it was trained on some legal documents from Indian government websites (such as State Assembly discussions) according to ~\citet{ramesh2022samanantar}. 
However, it is {\it not} known publicly over what data commercial systems such as GOOG and MSFT are trained.
By looking at the average scores across all 3 datasets and language pairs (see Table~\ref{tab:mt-allresults}), we can establish that MSFT performs the best across all metrics.}

\begin{table*}[t]
    \centering
    \small
    \renewcommand{\arraystretch}{0.7}
    \begin{tabular}{ccccccccc}
       \toprule
        \multirow{2}{*}{\textbf{Task}} & \multicolumn{3}{c}{\textbf{GPT-3.5}} & \multicolumn{3}{c}{\textbf{GPT-4}} & \multirow{2}{*}{\textbf{SOTA}} & \multirow{2}{*}{\textbf{Metric}} \\
        \cmidrule(lr){2-4} \cmidrule(lr){5-7}
        & \textbf{0-Shot} & \textbf{1-Shot} & \textbf{2-Shot} & \textbf{0-Shot} & \textbf{1-Shot} & \textbf{2-Shot} & & \\ \midrule
        
        \texttt{L-NER} & 30.59\% & 23.68\% & \uline{32.84}\%* & 13.65\% & 10.51\% & 24.03\% & \textbf{48.58}\% & strict mF1  \\ \midrule
         \texttt{RR} & 30.95\%  & 30.05\%  & 30.31\% & 37.37\% & 37.43\% & \underline{38.18}\% & \textbf{69.01}\% & mF1 \\ \midrule
         \texttt{CJPE} & \begin{tabular}{@{}c@{}}54.17\%\\ 0.30\\ 0.08\\ \end{tabular} & \begin{tabular}{@{}c@{}} 51.46\%\\ 0.29 \\ 0.15 \end{tabular} & \begin{tabular}{@{}c@{}} 56.74\% \\0.30 \\0.113 \end{tabular} & \begin{tabular}{@{}c@{}} \underline{68.29}\% \\0.40 \\0.14 \end{tabular} & \begin{tabular}{@{}c@{}} 47.26\% \\0.39 \\0.16 \end{tabular} & \begin{tabular}{@{}c@{}} 60.44\% \\\uline{0.43} \\\uline{0.18} \end{tabular} & \begin{tabular}{@{}c@{}} \textbf{81.31}\% \\ \textbf{0.56} \\ \textbf{0.32} \end{tabular} & \begin{tabular}{@{}c@{}} mF1 \\ ROUGE-L \\ BLEU \end{tabular} \\ \midrule
         \texttt{BAIL} &  51.04\%  & 46.35\%  & 61.0\% & 51.46\% & 56.90\% & \underline{66.67}\% & \textbf{81}\% & mF1 \\ \midrule
         \texttt{LSI} &  21.55\%  & 22.61\%  & 21.40\% & \uline{23.99} & 22.26 & 20.53 & \textbf{28.08}\% & mF1  \\ \midrule
         \texttt{SUMM} &  \begin{tabular}{@{}c@{}} 0.21 \\ \underline{0.85} \end{tabular} & \begin{tabular}{@{}c@{}} 0.20 \\ 0.84 \end{tabular} & \begin{tabular}{@{}c@{}}0.22 \\ 0.84 \end{tabular} & \begin{tabular}{@{}c@{}} \uline{0.23} \\ \uline{0.85} \end{tabular} & \begin{tabular}{@{}c@{}}0.16 \\ 0.81 \end{tabular} & \begin{tabular}{@{}c@{}}0.17 \\ 0.81 \end{tabular} & \begin{tabular}{@{}c@{}} \textbf{0.33} \\ \textbf{0.86} \end{tabular} & \begin{tabular}{@{}c@{}} ROUGE-L\\ BERTScore \end{tabular} \\ \midrule
         \texttt{L-MT} & \begin{tabular}{@{}c@{}} 0.23\\ 0.28 \\ 0.42 \end{tabular} & \begin{tabular}{@{}c@{}}0.25 \\0.28 \\ 0.43 \end{tabular} & \begin{tabular}{@{}c@{}} 0.26 \\ 0.29 \\ 0.43 \end{tabular} & \begin{tabular}{@{}c@{}} 0.33 \\ 0.36 \\ 0.50 \end{tabular} & \begin{tabular}{@{}c@{}} 0.35 \\ 0.38 \\ 0.52 \end{tabular} & \begin{tabular}{@{}c@{}} \uline{\textbf{0.36}} \\ \uline{\textbf{0.39}} \\ \uline{0.53} \end{tabular} &  \begin{tabular}{@{}c@{}} 0.28 \\ 0.32 \\ \textbf{0.57} \end{tabular} & \begin{tabular}{@{}c@{}} BLEU \\ GLEU \\ chrF++ \end{tabular} \\ 
         \bottomrule
    \end{tabular}
    \caption[All Tasks GPT Results]{Performance of Open-AI GPT-3.5 (\texttt{gpt-3.5-turbo-16k}) and GPT-4 (\texttt{gpt-4-turbo}) model on various tasks for zero-shot, one-shot and two-shot settings. The SOTA corresponds to the best performing model as given in Table \ref{tab:tasks-results}. The best result for each task is marked in \textbf{boldface}. The best GPT-based result for each task is \uline{underlined}.} 
    \label{app-tab:llm-results}
\end{table*}

\section{Additional Experiments with LLMs} \label{app-sec:llms}
The wide generalization capability of large language models has shown tremendous performance across various Natural Language Understanding (NLU) tasks. To validate if the available LLMs generalize enough to domain-specific legal language, we perform a detailed set of experiments by prompting LLMs over the set of proposed tasks in \benchmarkName. 
We design prompts based on the available task, the context length, and prior knowledge required for the task, like label definition, which is specific to the legal domain. 
In recent years, In-Context Learning (ICL) \cite{brown2020language-incontext-learning} has significantly improved LLMs performance on various tasks. Considering the performance boost due to the ICL prompt template, it becomes crucial to consider few-shot prompts. For our experiments with LLMs, we design a prompt template that is compatible with ICL, i.e., the same prompt template can be used to provide a few shot examples as a prompt to the language models. Primarily, we validate the performance of large proprietary LLMs as well as smaller non-commercial LLMs. 
As some of the tasks require the entire document to be a part of the model's input, evaluating the entire test sets becomes more challenging and time-consuming for tasks with large test sets. 
Since the primary design of the benchmark is not LLM specific, we perform the LLM validation to obtain a general proxy of LLM performance. 

\subsection{Experiments with Proprietary LLMs}
For experiments with proprietary LLMs, we consider the widely used models released by OpenAI: GPT-3.5 (\texttt{gpt-3.5-turbo-16k}) and GPT-4 (\texttt{gpt-4-turbo}). As explained in \S\ref{sec:models}, PCR requires as input the texts of the source document as well as a set of candidate documents. Due to the size of legal documents, such a setup exceeds the token length limit for GPT-3.5  and also for GPT-4 if all candidates are considered. Hence, we could not experiment with LLMs for the task of PCR. 
We discuss the task-specific prompt design and evaluation strategies and the obtained findings in the subsections below. Table \ref{app-tab:llm-results} shows the results for various tasks. 

\begin{table}[htbp]
    \tiny
    \centering
    \begin{tabular}{|p{0.98\linewidth}|}
    \hline
       \begin{alltt}
    \noindent \textbf{SYSTEM\_PROMPT:} You are a smart and intelligent Named Entity Recognition (NER) system. I will provide you with the definition of the entities you need to extract and the output format. I will also provide some examples of the task and the document from where you should extract the entities.
    
    \noindent \textbf{USER\_PROMPT:} Are you clear about your role?

    \noindent \textbf{ASSISTANT\_PROMPT:} Sure, I'm ready to help you with your NER task. Please provide me with the necessary information to get started.

    \noindent \textbf{INPUT\_PROMPT:} 

    \noindent \textbf{Entity Definition:}
    
    1. APPELLANT: Name or abbreviation of the person(s) or organization(s) filing an appeal/petition to a court of law.
    
    2. RESPONDENT: Name or abbreviation of a person(s) or organization(s) responding/defending to an appeal/petition filed against them in a court of law.
    
    3. JUDGE: Name of the judge/justice presiding over the case in a court of law.
    
    4. APPELLANT COUNSEL: Name of the lawyer representing the appellant/petitioner in a court of law.
    
    5. RESPONDENT COUNSEL: Name of the lawyer representing the respondent in a court of law.
    
    6. COURT: Name of the court of law
    
    7. AUTHORITY: Name or abbreviation of any organization apart from a Court, which has administrative, legal or financial authority. This also includes regulatory and investigative agencies.
    
    8. WITNESS: Name of a person appearing as witness or testifying to a case in a court of law.
    
    9. STATUTE: Name or abbreviation of a statutory law or legal article.
    
    10. PRECEDENT: Title of a prior court case.
    
    11. DATE: Any format of date, even in natural language.
    
    12. CASE NUMBER: Any format of prior case number or order numbers.
    
    Important Instructions:
    
    1. Salutations or prefixes/suffixes like Mr., Mrs., Smt., Justice, J., Dr., P.W., are not part of the named entity.

    \noindent \textbf{Output Format:}
    
    \noindent\{"APPELLANT": [list of entities present], "RESPONDENT": [list of entities present], "JUDGE": [list of entities present], "APPELLANT COUNSEL": [list of entities present], "RESPONDENT COUNSEL": [list of entities present], "COURT": [list of entities present], "AUTHORITY": [list of entities present], "WITNESS": [list of entities present], "STATUTE": [list of entities present], "PRECEDENT": [list of entities present], "DATE": [list of entities present], "CASE NUMBER": [list of entities present]\}
    
    DO NOT REPEAT THE SAME ENTITY NAME MULTIPLE TIMES.
    
    If no entities are presented in any category, keep an empty list for that category.

    The above format should be a pure JSON format.

    \textbf{\noindent Examples:}

    \textbf{Document 1: \textit{<In-context Document 1 goes here>}}

    \textbf{Output 1: \textit{<Gold-standard Labels for Document 1 goes here>}}

    \textbf{\ldots}

    \textbf{Document \textit{n+1}: \textit{<Test Document goes here>}}

    \textbf{Output \textit{n+1}:}
\end{alltt} \\\hline
    \end{tabular}
    \caption[L-NER Task GPT Prompt]{Prompt template for L-NER for both GPT-3.5 and GPT-4 ($n$ in-context examples)}
    \label{prompt:lner}
\end{table}

\subsubsection{Legal Named Entity Recognition (L-NER)}
\label{app:lner-chatgpt}


\noindent\textbf{Prompt Design:}
Although the NER task is known to GPT, LNER involves clearly understanding the meaning of the legal entities. Thus, we provide descriptions of the entities as part of our prompt (Table~\ref{prompt:lner}).

\noindent\textbf{Data Selection:}
As discussed in App.~\ref{app:lner-models}, we divided our entire data into 3 folds for testing the other models. In this experiment, we only choose the documents of one particular fold (Fold 1) for passing to GPT.
For in-context learning, we randomly sample documents from Fold 2. In some cases, especially for 2-shot prompting, the input did not fit within 16k tokens (for GPT-3.5) even after choosing the shortest in-context (IC) examples.
In these cases, we split the document into chunks, passed each chunk to the model along with IC examples, and collated the outputs from each chunk to produce the final output. 
No such adjustments were needed for GPT-4 due to its larger context length (128k tokens).

\noindent\textbf{Verbalization:} We expect the model's output to be precisely compatible with JSON. For GPT-3.5, the generated JSON format was sometimes incomplete, and we used string processing to complete these strings for JSON compatibility. For GPT-4, all results were perfectly JSON compatible.

\begin{table}
    \centering
    \tiny
    \renewcommand{\arraystretch}{1.3}
    \setlength\tabcolsep{4pt}
    \hspace{0.2cm}
    \begin{tabular}{l | c  c  c | c  c  c}
       \hline
        \multirow{2}{*}{\textbf{Method}} & \multicolumn{3}{c|}{\textbf{Strict}}& \multicolumn{3}{c}{\textbf{Ent type}}\\
        & \textbf{mP} & \textbf{mR} & \textbf{mF1} & \textbf{mP} & \textbf{mR} & \textbf{mF1}\\ \hline
        GPT-3.5 0-shot& 48.57& 24.58& 30.59& 65.23& \textbf{34.46}& \textbf{42.04}\\ 
        GPT-3.5 1-shot& 39.08& 18.56& 23.68& 56.05& 26.73&34.34\\
        GPT-3.5 2-shot& \textbf{51.29}& \textbf{26.16}& \textbf{32.84}& \textbf{65.63}& 32.80&41.54\\\hline
        GPT-4 0-shot& 21.44& 10.10& 13.65& 22.62& 10.64& 14.37\\ 
        GPT-4 1-shot& 20.32& 7.49& 10.51& 22.69& 8.34&11.72\\
        GPT-4 2-shot& 47.30& 18.26& 24.03& 53.50& 20.82&27.30\\\hline
    \end{tabular}
    \caption[L-NER Task GPT Results]{Performance of GPT-3.5 and GPT-4 over the L-NER dataset. All values are macro-averaged and in terms of percentage.}
    \label{tab:ner-cgpt-results}
\end{table}

\noindent\textbf{Results:}
GPT returns a list of entities for each class. We mapped all character spans in the document corresponding to each entity and used these character span mappings to generate the BIO sequence that is further used for evaluation. The results for the GPT are mentioned in Table~\ref{tab:ner-cgpt-results}. Firstly, we observe that GPT-4 performs very poorly as compared to GPT-3.5 (discussed below). In terms of the strict scores, GPT-3.5 performs much poorly compared to the SOTA models, demonstrating that it cannot understand the legal roles clearly without any fine-tuning. Observing the 1 and 2-shot results, it is clear that providing a single ICL example can mislead the model, and adding 2 examples provides a slight improvement over 0-shot. Finally, as observed for the BERT-based models, there is a significant difference between strict and ent-type scores.

The massive drop in performance for GPT-4 requires further investigation. We experimented by lowering the temperature, but this led to even worse performance. Similarly, we tried to modify the prompt to make the model focus on covering all entities, variations, etc. But none of these techniques improved the performance. We manually verified the outputs of GPT-3.5 and GPT-4 in comparison with the gold standard. We observed that GPT-4 was hallucinating to a great extent, returning many new entities that are not present in the input document. It also failed to capture many true entities in the process, which explains the poor precision and recall values. To further highlight the above issue, we show the predictions of GPT-3.5 and GPT-4 in comparison with the gold-standard entities for a particular case ``Babulal Badriprasad Varma vs Surat Municipal Corpn. \& Ors'', Supreme Court of India (2008), in Table~\ref{tab:lner_gpt_outputs}. Entities not present in the case document are marked in red. We can clearly see that GPT-4 produces hallucinated outputs across all labels. While it can correctly pick up true entities as well, a large number of these hallucinated outputs lead to very poor metrics. On further inspection, we observed that all these hallucinated entities came from the same case, ``Sonali Hatua Giri vs Union Of India And Others on 7 April, 2021'', Calcutta High Court (Appellate Side) (2021).

\begin{table*}
    \centering
    \tiny
    \begin{tabular}{p{0.15\textwidth}p{0.25\textwidth}p{0.25\textwidth}p{0.25\textwidth}}
        \toprule
         \textbf{Class}&  \textbf{Gold Standard}&  \textbf{GPT-3.5}& \textbf{GPT-4}\\ \midrule
         APP & {Babulal Badriprasad Varma}  & Babulal Badriprasad Varma  & \textcolor{red}{Sonali Hatua Giri}, Babulal Badriprasad Varma \\ \midrule
         RESP & Surat Municipal Corporation & Surat Municipal Corporation & \textcolor{red}{Union of India}, Surat Municipal Corporation \& Ors.\\ \midrule
         JUDGE & V.S. Sirpurkar, S.B. Sinha, Johan Romilly M. R.  & S.B. Sinha & \textcolor{red}{Sabyasachi Bhattacharyya}, S.B. Sinha \\ \midrule
         A.COUNSEL & Lalit, U.U. Lalit & Mr. U.U. Lalit & \textcolor{red}{Sankar Nath Mukherjee}, \textcolor{red}{Niraj Gupta}, \textcolor{red}{Priyabrata Shah}, \textcolor{red}{Aditya Biswas}, \textcolor{red}{U.U. Lalit}\\ \midrule
         R.COUNSEL & Prashant G. Desai & Mr. Prashant G. Desai & \textcolor{red}{Y. J. Dastoor}, \textcolor{red}{Rudraman Bhattacharya}, \textcolor{red}{Anuran Samanta}, Prashant G. Desai \\ \midrule
         WITNESS & & & \\ \midrule
         COURT & Supreme Court of India, Criminal Court, Apex Court, High Court of Gujarat, High Court & Supreme Court of India & \textcolor{red}{Punjab and Haryana High Court}, Supreme Court, High Court of Gujarat \\ \midrule
         AUTHORITY & Government of Gujarat, Parliament & Government of Gujarat, High Court of Gujarat, Municipal Corporation & \textcolor{red}{Ministry of Home Affairs}, \textcolor{red}{Central/State Government}, Government of Gujarat \\ \midrule
         STATUTE & Gujarat Town Planning and Urban Development Rules, 1979; Transfer of Property Act; Bombay Town Planning Rules, 1955; Real Property Act; Bangalore Development Authority Act, 1976; Indian Penal Code; Bombay Provincial Municipalities Act; Gujarat Town Planning and Urban Development Act & Gujarat Town Planning and Urban Development Act, 1976; Gujarat Town Planning and Urban Development Rules, 1979; Bombay Provincial Municipalities Act; Indian Penal Code & \textcolor{red}{Central Samman Pension Scheme}; Constitution of India; \textcolor{red}{Hindu Succession Act, 1956}; \textcolor{red}{Indian Succession Act, 1925}; \textcolor{red}{Hindu Marriage Act}; \textcolor{red}{Indian Divorce Act, 1969}; \textcolor{red}{Muslim Women (Protection of Rights on Divorce) Act, 1986}; \textcolor{red}{Parsi Marriage and Divorce Act, 1936}; Gujarat Town Planning and Urban Development Act, 1976; Gujarat Town Planning and Urban Development Rules, 1979 \\ \midrule
         PRECEDENT & Maneklal Chhotalal \& Ors. v. M.G. Makwana \& Ors. [(1967) 3 SCR 65]; Maneklal Chhotalal (supra); Bhikhubhai Vithlabhai Patel \& Ors. v. State of Gujarat \& Anr. 2008 (4) SCALE 278; Manak Lal v. Dr. Prem Chand [AIR 1957 SC 425]; Director of Inspection of Income Tax (Investigation), New Delhi and Another v. Pooran Mal \& Sons and Another [(1975) 4 SCC 568]; Bank of India v. O.P. Swarnakar (2003) 2 SCC 721, Mansukhlal (supra); Jaswantsingh Mathurasingh and Another v. Ahmedabad Municipal Corporation and Others [1992 Supp (1) SCC 5]; State of Gujarat v. Shantilal Mangaldas \& Ors. [1969 (3) SCR 341]; Sureshchandra C. Mehta v. State of Karnataka and Others 1994 Supp (2) SCC 511; Mansukhlal Jadavji Darji (supra); West Bengal Housing Board etc. v. Brijendra Prasad Gupta and Others, etc. [AIR 1997 SC 2745]; Phillips v. Martin; Mansukhlal Jadavji Darji and Others v. Ahmedabad Municipal Corporation and Others [(1992) 1 SCC 384]; Vyvyan v. Vyvyan [(1861) 30 Beav. 65, 74; 54 E.R. 813, 817]; Jaswantsingh Mathurasingh (supra); Ramdev Food Products Pvt. Ltd. v. Arvindbhai Rambhai Patel and Ors. [2006 (8) SCALE 631]; Krishna Bahadur v. Purna Theatre [(2004) 8 SCC 229]; Wilson v. McIntosh & Mansukhlal Jadavji Darji v. Ahmedabad Municipal Corporation; Jaswantsingh Mathurasingh v. Ahmedabad Municipal Corporation; State of Gujarat v. Shantilal Mangaldas \& Ors. & \textcolor{red}{Municipal committee Patiala Vs. Model town Residents Association}; \textcolor{red}{Khajani Devi Vs. Union of India and others}; \textcolor{red}{Tulsi Devi Vs. Union of India and another}; Mansukhlal Jadavji Darji and Others v. Ahmedabad Municipal Corporation and Others; Jaswantsingh Mathurasingh and Another v. Ahmedabad Municipal Corporation and Others \\ \midrule
         DATE & 27.12.2006; 1.06.1999; 23.11.2006; May 02, 2008; 31.03.2000; 15.01.2000; 1.7.1999; 15.1.2000 & 27.12.2006; 23.11.2006; 1.06.1999; 15.01.2000; 31.03.2000 & \textcolor{red}{07.04.2021}; \textcolor{red}{December 4, 2012}; \textcolor{red}{March 19, 1999}; \textcolor{red}{February 18, 2019}; \textcolor{red}{July 29, 2016}; \textcolor{red}{September 27, 2019}; \textcolor{red}{July 18, 2019}; \textcolor{red}{May 28, 2020}; 27.12.2006, 23.11.2006, 1.06.1999, 15.01.2000, 31.03.2000 \\ \midrule
         CASE NO. & CIVIL APPEAL NO. 3203 OF 2008; SCA No. 7092 of 2001; Letters Patent Appeal No. 1611 of 2006; SLP (Civil) No. 568 of 2007 & CIVIL APPEAL NO. 3203 OF 2008; SLP (Civil) No. 568 of 2007; Letters Patent Appeal No. 1611 of 2006; SCA No. 7092 of 2001 & \textcolor{red}{WPA 13806 of 2019}; \textcolor{red}{CWP No.1504 of 2019}; \textcolor{red}{No.17706 of 2017}; CIVIL APPEAL NO. 3203 OF 2008, SLP (Civil) No. 568 of 2007, Letters Patent Appeal No. 1611 of 2006, SCA No. 7092 of 2001 \\ \bottomrule
         
    \end{tabular}
    \caption[L-NER Task GPT Outputs]{Predictions of GPT-3.5 and GPT-4 for the LNER task, compared with the gold standard. Entities not present in the original case document are marked in \textcolor{red}{red}.}
    \label{tab:lner_gpt_outputs}
\end{table*}


\subsubsection{Rhetorical Role Labeling (RR)}\label{app:rr-chatgpt}

\noindent\textbf{Prompt Design:}
{The RR task can be considered a semantic role labeling task over the sentences. Such a variant of the task and the definition of the rhetorical roles themselves are probably not clearly known to the GPT models; hence, we give explicit guidelines on how to carry out the labeling task.
We tried out some initial prompts considering document-level inputs, i.e., passing the entire document (list of sentences) to GPT-3.5 and asking it to generate a list of labels corresponding to each sentence. This approach had several challenges, such as the output not having the same number of labels as input sentences, random token generation, etc.
This problem became more pronounced in the ICL setting.
Further, input text and sample output for IC examples were becoming too long.
Thus, for both GPT-3.5 and GPT-4, we frame the task as a simple sentence classification task, asking the models to predict the label of an individual sentence.
The final prompt is shown in Table~\ref{prompt:rr}.
We run both GPT-3.5 and GPT-4 over all sentences in a document to get all corresponding label predictions.
}

\begin{table}[htbp]
    \tiny
    \centering
    \begin{tabular}{|p{0.9\linewidth}|}\hline
        
       \begin{alltt}
    \noindent \textbf{SYSTEM\_PROMPT:} You are a smart and intelligent legal semantic role labeling system. In Indian Court judgment documents, each document sentence can be assigned a legal semantic role. Your task is, given a sentence from an Indian Court case document, to identify the given sentence's semantic role. I will provide you with the descriptions of the legal semantic roles. I will also provide you with some examples.

    \vspace{1mm}
    \noindent \textbf{USER\_PROMPT:} Are you clear about your role?

    \vspace{1mm}
    \noindent \textbf{ASSISTANT\_PROMPT:} Absolutely, I understand my role. You would like me to identify a sentence's legal semantic role label in an Indian court case document. Please provide me with the descriptions of the legal semantic roles to help guide me in accurately assigning the role to the given sentence.

    \vspace{1mm}
    \noindent \textbf{INPUT\_PROMPT: }

    \noindent Legal Semantic Role Descriptions:
    
    1. Fact: The actual facts and events that led to the case.
    
    2. Argument: Legal arguments which have been put forward by either lawyer.

    3. RulingByLowerCourt: Decisions of the lower courts, if any.
    
    4. Statute: References or citations to statutory laws and articles referred in the case.
    
    5. Precedent: Sentences containing References or citations to precedents (prior cases). 
    
    6. RatioOfTheDecision: The reasoning which has been established by the judge in the current judgment.
        
    7. RulingByPresentCourt: The final decision of the current court.
    
    \noindent ANSWER ONLY WITH ONE OF THE ABOVE CHOICES, DO NOT PROVIDE ANY EXTRA OUTPUT.
    
    \noindent \textbf{Examples:}
    
    \textbf{Sentence 1: <In-context Sentence 1 goes here>}

    \textbf{Output 1: <Gold-standard Label for Sentence 1 goes here>}

    \textbf{\ldots}

    \textbf{Sentence n+1: <Test Sentence goes here>}

    \textbf{Output n+1:}        
\end{alltt} \\\hline
    \end{tabular}
    \caption[RR Task GPT Prompt]{Prompt template for RR (for $n$ in-context examples) for both GPT-3.5 and GPT-4.}
    \label{prompt:rr}
\end{table}


\noindent\textbf{Data Selection:}
{We used all sentences from all documents in the CL and IT test sets (5 documents each). For in-context samples, we randomly choose sentences from all these documents except the document from which the test sentence (sentence for which we expect GPT prediction) is sampled.}

\begin{table}
    \centering
    \small
    \begin{tabular}{cccc}\toprule
         \textbf{Model}&  \textbf{CL} &  \textbf{IT}& \textbf{CL + IT}\\\midrule
         GPT-3.5 0-shot&  0.25&  0.37& 0.31\\
         GPT-3.5 1-shot&  0.24&  0.36& 0.30\\
         GPT-3.5 2-shot&  0.23& 0.38 & 0.30\\\midrule
         GPT-4 0-shot&  \textbf{0.29} &  0.46& 0.37\\
         GPT-4 1-shot&  \textbf{0.29}&  0.45& 0.37\\
         GPT-4 2-shot&  0.28&  \textbf{0.49}& \textbf{0.38}\\\bottomrule
    \end{tabular}
    \caption[RR Task GPT Results]{Macro-F1 scores for RR datasets}
    \label{tab:rr_results}
\end{table}

\noindent\textbf{Verbalizer:} In most cases, GPT-3.5 answers with the exact label name. In some cases, it can be accompanied by extra erroneous words. In case the prediction is a sequence of words, we iterate over the words and choose the first word that corresponds to an RR. If no such word is found, GPT-3.5 prediction has failed, and we randomly choose a label to substitute its decision. We did not observe such anomalies for GPT-4.

\noindent\textbf{Results:}
{The SOTA model achieves a macro-F1 of 70\% over the combined (IT + CL) test set. In comparison, GPT-3.5 can only achieve a macro-F1 of 31\%, showing that it is not straightforward for the LLM to assign semantic labels to sentences.
On manual inspection, we observed that the model was prone to assign the FAC label to all sentences with the model temperature set to 0.
On increasing the temperature to 0.95 (temperature 1 was not giving stable results), we observe that the model is still prone to assigning labels like FAC, ARG-P, ARG-R, and RPC (frequent labels) to most sentences. GPT-4 consistently outperforms GPT-3.5, with the improvements being more significant for IT documents. Also, it seems that ICL has no positive impact on either GPT-3.5 or GPT-4, with there being minimal or no improvements at all. It could be possible that just the description of the labels is not enough; GPT models might need 1/2 examples from each class to clearly understand the meaning of the RRs. However, this approach is likely to increase the context length significantly.
}



\subsubsection{Court Judgment Prediction and Explanation (CJPE)}
\label{app:cjpe-chatgpt}

\noindent\textbf{Prompt Design:}
{For the prediction aspect of this task, we ask GPT to read the content of the entire document and predict the final ``accept''/``reject'' decision (Table~\ref{prompt:cjpe-pred}). For the explanation aspect, we modify the prompt, asking GPT first to predict the accept/reject decision and then extract important sentences of the text that led to its decision (Table~\ref{prompt:cjpe-exp}).} While the exact same prompt is used for both GPT-3.5 and GPT-4 for the classification task, slightly different prompts were used for the explainability task for each model.

\begin{table}[htbp]
    \tiny
    \centering
    \begin{tabular}{|p{0.9\linewidth}|}\hline
        
       \begin{alltt}

    \vspace{1mm}
    \noindent \textbf{SYSTEM\_PROMPT:} You are a smart and intelligent system, trained to act like a judge in the Indian Supreme Court. A court case document in the Indian Supreme Court can consist of one or more appeals by a particular party. Your task is, given such a case document, to predict whether the appeals will be accepted or rejected. For cases containing multiple appeals, you will predict either 'accept' if at least one of the appeals can be accepted or 'reject' if none of the appeals can be accepted. PLEASE ANSWER ONLY WITH EITHER 'ACCEPT' OR 'REJECT'. I will provide you with some examples of this task and the case document you need to make the prediction for.
    
          \vspace{1mm}
    \noindent \textbf{USER\_PROMPT:} Are you clear about your role?

    \vspace{1mm}
    \noindent \textbf{ASSISTANT\_PROMPT: }Sure, I'm ready to help you with your court judgment prediction task. Please provide me with the examples and the case document I'm supposed to make the prediction for.

    \vspace{1mm}
    \noindent \textbf{INPUT\_PROMPT: }

    \noindent \textbf{Examples:}
    
    \textbf{Case Document 1: <In-context Document 1 goes here>}

    \textbf{Output 1: <Gold-standard Label for Document 1 goes here>}

    \textbf{\ldots}

    \textbf{Case Document n+1: <Test Document goes here>}
    
    \textbf{Output:}  
\end{alltt} \\\hline
    \end{tabular}
    \caption[CJPE Prediction Task GPT Prompt]{Prompt template for CJPE Prediction for both GPT-3.5 and GPT-4 ($n$ in-context examples)}
    \label{prompt:cjpe-pred}
\end{table}

\noindent\textbf{Data Selection:}
For prediction, we divide the \texttt{ILDC-multi} test set into positive and negative examples and randomly sample 50 positive and 50 negative examples. For ICL, we randomly sample examples from the remaining test set documents such that the final prompt is within the GPT token limit.
For explanation, we use all 56 documents from \texttt{ILDC-expert} for prompting. We sample the IC examples from this set itself. 
The gold standard outputs, in this case, are the important sentences with rank 1 and 2, as per the ranking given by either expert 3 or expert 4, chosen randomly (since these experts had the highest agreement according to \citet{malik2021ildc}).
In both cases, for 2-shot prompting, we sample one document each from the positive and negative classes.

\noindent\textbf{Verbalizer:}
For classification, the model always answers with either ACCEPT/REJECT. For explanation, in the case of GPT-3.5, we did not issue strict output format instructions since GPT-3.5 was unable to understand them correctly (based on few documents of the validation set). Thus, there were a few variations in the output format, but included a list of the important sentences, marked either with bullet points, numbering, or other delimiters. We used these cues to extract the exact sentences. However, GPT-4 can understand and adhere to complex instructions much better and thus we could specify stricter rules regarding the output format, making the task of verbalization easier for GPT-4.

\begin{table}[htbp]
    \tiny
    \centering
    \begin{tabular}{|p{0.9\linewidth}|}\hline
        
       \begin{alltt}

    \vspace{1mm}
    \noindent \textbf{SYSTEM\_PROMPT:} You are a smart and intelligent system, trained to act like a judge in the Indian Supreme Court. A court case document in the Indian Supreme Court can consist of one or more appeals by a particular party. Your task is, given such a case document, to predict whether the appeals will be ACCEPTED or REJECTED. You will also have to explain your prediction by QUOTING VERBATIM the important sentences of the input text that led to your decision. I will provide you with examples of the task, and then the input document.
    
          \vspace{1mm}
    \noindent \textbf{USER\_PROMPT:} Are you clear about your role?

    \vspace{1mm}
    \noindent \textbf{ASSISTANT\_PROMPT}: Sure, I'm ready to help you with your court judgment prediction task. I will also quote verbatim the important areas of the input text that led to my prediction. Please provide me with the case document I'm supposed to make the prediction for.

    \vspace{1mm}
    \noindent \textbf{INPUT\_PROMPT:} 

    \noindent \textbf{IMPORTANT:} For explaining your prediction, quote important sentences verbatim from the input text. DO NOT PARAPHRASE OR SUMMARIZE THESE SENTENCES.

    \noindent \textbf{Examples:}
    
    \textbf{Case Document 1: <In-context Document 1 goes here>}

    \textbf{Output 1:} The appeals will be \textbf{<Gold-standard ACCEPT/REJECT Label for Document 1 goes here>}. Here are the verbatim sentences that led to this decision: \textbf{<Gold-standard important sentences for Document 1 goes here>}

    \textbf{\ldots}

    \textbf{Case Document n+1: <Test Document goes here>}
    
    \textbf{Output:}  
\end{alltt} \\\hline
    \end{tabular}
    \caption[CJPE Explanation Task GPT-3.5 Prompt]{Prompt template for CJPE Explantion for GPT-3.5 ($n$ in-context examples)}
    \label{prompt:cjpe-exp}
\end{table}

\begin{table}[htbp]
    \tiny
    \centering
    \begin{tabular}{|p{0.9\linewidth}|}\hline
        
       \begin{alltt}

    \vspace{1mm}
    \noindent \textbf{SYSTEM\_PROMPT:} You are a smart and intelligent system, trained to act like a judge in the Indian Supreme Court. A court case document in the Indian Supreme Court can consist of one or more appeals by a particular party. Your task is, given such a case document from the Indian Supreme Court, predict whether the appeals will be ACCEPTED or REJECTED. You will also have to explain your prediction by printing the important sentences of the input text that primarily influenced your decision. Make sure to include any sentence which is even slightly relevant for your final decision. Do not paraphrase, summarize or change the sentences in any way while printing, you must quote the sentences verbatim from the input case document. I will provide you with the output format and the input case document. I will also provide some examples of the task to help you learn from.
    
          \vspace{1mm}
    \noindent \textbf{USER\_PROMPT:} Are you clear about your role?

    \vspace{1mm}
    \noindent \textbf{ASSISTANT\_PROMPT}: Sure, I'm ready to help you with your court judgment prediction task. I will also quote verbatim the important sentences of the input text that led to my prediction. Please provide me with the example and the case document.

    \vspace{1mm}
    \noindent \textbf{INPUT\_PROMPT:} 

    \noindent \textbf{IMPORTANT:} Strictly adhere to the output format given below. Quote the important sentences verbatim from the input document.

    \noindent \textbf{Output Format:} 

    \noindent DECISION: ACCEPTED/REJECTED

    \noindent Important Sentences:

    1. First important sentence

    2. Second important sentence

    \ldots

    \noindent \textbf{Examples:}
    
    \textbf{Case Document 1: <In-context Document 1 goes here>}

    \textbf{Output 1:}

    DECISION: \textbf{<Gold-standard ACCEPT/REJECT Label for Document 1 goes here>}

    Important Sentences:

    1. \textbf{<Gold-standard First Important Sentence of Document 1 goes here>}

    2. \textbf{<Gold-standard Second Important Sentence of Document 1 goes here>}

    \textbf{\ldots}

    \textbf{Case Document n+1: <Test Document goes here>}
    
    \textbf{Output n+1:}  
\end{alltt} \\\hline
    \end{tabular}
    \caption[CJPE Explanation Task GPT-4 Prompt]{Prompt template for CJPE Explanation for GPT-4 ($n$ in-context examples)}
    \label{prompt:cjpe-exp}
\end{table}

\begin{table}[htbp]
    \centering
    \footnotesize
    \renewcommand{\arraystretch}{1.3}
    \setlength\tabcolsep{3pt}
    \hspace{0.2cm}
    \begin{tabular}{p{0.28\columnwidth}  c  c  c  c  c  c}
       \toprule
        \multirow{2}{*}{\textbf{Model}} & \multicolumn{3}{c}
        {$\text{ILDC}_{\text{multi}}$}& \multicolumn{2}{c}{$\text{ILDC}_{\text{expert}}$}\\
        & \textbf{mP} & \textbf{mR} & \textbf{mF1} & \textbf{R-L} & \textbf{BLEU}\\ \bottomrule
        GPT-3.5 0-shot& 57.14& 56.00& 54.17& 0.301& 0.077\\ 
        GPT-3.5 1-shot& 57.06& 55.00& 51.46& 0.292& 0.155\\
        GPT-3.5 2-shot& 65.79& 60.42& 56.74& 0.299& 0.113\\\midrule
        GPT-4 0-shot& 70.88& \textbf{69.00} & \textbf{68.29}& 0.398& 0.137\\ 
        GPT-4 1-shot& 55.31& 53.00& 47.26& 0.395& 0.161\\
        GPT-4 2-shot& \textbf{71.88}& 64.00& 60.44& \textbf{0.426}& \textbf{0.184}\\\bottomrule
        
    \end{tabular}
    \caption[CJPE Task GPT Results]{Performance over the CJPE datatsets. P, R and F1 values are macro-averaged and in terms of percentage.}
    \label{tab:cjpe-cgpt-results}
\end{table}

\noindent\textbf{Results:}
For prediction, both GPT-3.5 and GPT-4 tend to predict ``reject'' in favor of ``accept'' in most cases.
Only by tweaking the temperature up to as high as $0.98$, we could observe more ``accept'' predictions.
Despite this, GPT-3.5 significantly underperforms compared to SOTA approaches, barely performing better than random choice (see Table~\ref{tab:cjpe-cgpt-results}). 
The result turns even worse with 1-shot prompting, possibly making the model biased towards the class of the IC example. 2-shot prompting gives the best result among these settings for GPT-3.5. GPT-4, however, produces quite decent performance in 0-shot setting. It seems that the addition of ICL  examples is greatly detrimental to GPT-4, producing the worst performance in 1-shot setting, while also showing a drastic decrease in performance for the 2-shot setting.

The explanation is a more difficult task than the prediction, and GPT-3.5 again underperforms compared to the SOTA approach, especially considering the BLEU score (Table~\ref{tab:cjpe-cgpt-results}). ICL does not change the R-L score but has a positive impact on the BLEU score in the 1-shot setting only. GPT-4 shows significant improvement over GPT-3.5, both in terms of R-L and BLEU scores. ICL improves the R-L score under 2-shot setting only, while BLEU score improves progressively from 0-shot to 2-shot. The superior understanding capability of GPT-4 helps it perform better for both prediction and explanation, as compared to GPT-3.5.




\subsubsection{Bail Prediction (BAIL)}

\label{app:bail-chatgpt}

\noindent\textbf{Prompt Design:}
{BAIL is a binary classification task, and in terms of understanding and format, it is very similar to the CJPE task, the only difference being that the HLDC dataset for BAIL contains Hindi text rather than English. We use the same prompt for both GPT-3.5 and GPT-4, asking the models to read the application's content and provide the final decision, i.e., if the bail will be granted or dismissed (see Table~\ref{prompt:bail}).}

\begin{table}[htbp]
    \tiny
    \centering
    \begin{tabular}{|p{0.9\linewidth}|}\hline
        
       \begin{alltt}

    \vspace{1mm}
    \noindent \textbf{SYSTEM\_PROMPT:} You are a smart and intelligent system, trained to act like a judge in a district court of India. Most criminal cases in district courts involve bail applications written in Hindi. The application can be 'granted' if the judge believes the applicant deserves relief or 'dismissed' if the crime is too grave to grant relief. Your task is, given such a bail application, to predict if the bail will be 'granted' or 'dismissed'. PLEASE ANSWER ONLY WITH EITHER 'GRANTED' OR 'DISMISSED'. I will provide you with some examples of this task and the application document you need to make the prediction for.
    
          \vspace{1mm}
    \noindent \textbf{USER\_PROMPT:} Are you clear about your role?

    \vspace{1mm}
    \noindent \textbf{ASSISTANT\_PROMPT: }Sure, I'm ready to help you with your bail application prediction task. Please provide me with the examples and the bail application I'm supposed to make the prediction for.

    \vspace{1mm}
    \noindent \textbf{INPUT\_PROMPT: }

    \noindent \textbf{Examples:}
    
    \textbf{Bail Application 1: <In-context Application 1 goes here>}

    \textbf{Output 1: <Gold-standard Label for Application 1 goes here>}

    \textbf{\ldots}

    \textbf{Bail Application n+1: <Test Application goes here>}
    
    \textbf{Output:}  
\end{alltt} \\\hline
    \end{tabular}
    \caption[BAIL Task GPT Prompt]{Prompt template for BAIL Prediction for both GPT-3.5 and GPT-4 ($n$ in-context examples)}
    \label{prompt:bail}
\end{table}

\noindent\textbf{Data Selection:}
{We divide the \texttt{HLDC-all-districts} test set into positive and negative examples and randomly sample 50 positive and 50 negative examples that can be accommodated in the token length limit.
For ICL, we sample examples at random from the rest of the test set. For the 2-shot setting, we always sample one example each from the positive and negative classes.}

\noindent\textbf{Verbalizer:} Both GPT-3.5 and GPT-4 outputs only GRANTED/DISMISSED, so we directly take the model output as the predicted label.

\begin{table}
    \centering
    \small
    \begin{tabular}{cccc}\toprule
         \textbf{Model}&  \textbf{mP} &  \textbf{mR}& \textbf{mF1}\\ \midrule
         GPT-3.5 0-shot&  52.22&  52.00& 50.74\\
         GPT-3.5 1-shot&  46.85&  47.00& 46.35\\
         GPT-3.5 2-shot&  63.37&  62.00& 61.00\\ \midrule
         GPT-4 0-shot&  57.06&  55.00& 51.46\\
         GPT-4 1-shot&  58.91&  58.00& 56.90\\
         GPT-4 2-shot&  \textbf{71.43}&  \textbf{68.00}& \textbf{66.67}\\ \bottomrule
    \end{tabular}
    \caption[BAIL Task GPT Results]{Performance over the HLDC dataset for BAIL. P, R and F1 values are macro-averaged and in terms of percentage.}
    \label{tab:bail_cgpt_results}
\end{table}

\noindent\textbf{Results:}
Both GPT-3.5 and GPT-4 perform poorly in the 0-shot setting. As with CJPE, we observed a much higher proportion of negative class predictions. However, unlike CJPE, adjusting the temperature did not help too much. 1-shot ICL degrades the performance for GPT-3.5, possibly biasing the model to the class of the ICL example. GPT-4 is able to overcome that bias, showing improvements. Both models perform significantly better with 2 ICL examples, with GPT-4 performing the best by a margin.



\subsubsection{Legal Statute Identification (LSI)}
\label{app:lsi-chatgpt}

\noindent\textbf{Prompt Design:}
{The Indian Penal Code (IPC) is already known to GPT models since both GPT-3.5 and GPT-4 can accurately answer when asked about the content of different Sections of IPC.
In an initial setting, we asked both GPT-3.5 and GPT-4 to just output the list of relevant Section numbers of IPC for a given input.
We observed that the models produced hallucinated or completely non-relevant outputs; in this case, the output for GPT-3.5 often consisted of \textit{non-existent} IPC Section numbers.
Now, each IPC Section contains a corresponding title, which is a very short description of the entire statute.
In a second setting, we asked both models to output the section numbers and their corresponding titles.
For instance, if, for a particular case, Section 302 of the IPC is relevant, the model was expected to output just ``Section 302'' in the first setting, whereas it was expected to answer ``Section 302 --- Punishment for murder'' in the second setting.
We observed that this second setting reduced the hallucination to a great extent and improved performance. 
We did not specify any strict output format for GPT-3.5 since it was causing wrong predictions. However, we used this strategy for GPT-4, asking the model to print the relevant statutes in a structured format.
The prompts are shown in Table~\ref{prompt:lsi-3.5} (GPT-3.5) and Table~\ref{prompt:lsi-4}.}

\begin{table}[htbp]
    \tiny
    \centering
    \begin{tabular}{|p{0.9\linewidth}|}\hline
        
       \begin{alltt}

    \vspace{1mm}
    \noindent \textbf{SYSTEM\_PROMPT:} You are an intelligent Legal Crime Classification system. In the Indian legal system, the Indian Penal Code (IPC) is an Act in the Indian legislature that contains many legal articles or 'Sections' that codify different laws. Your task is, given the facts or evidence of an Indian court case as input, to predict the relevant or violated 'Sections' of the IPC as output. I will provide you some examples of this task and the facts of the case to make predictions for.
    
          \vspace{1mm}
    \noindent \textbf{USER\_PROMPT:} Are you clear about your role?

    \vspace{1mm}
    \noindent \textbf{ASSISTANT\_PROMPT: }Yes, I understand my role as an intelligent Legal Crime Classification system for the Indian legal system. You can provide me with the facts of a court case, and I will identify the relevant or violated sections of the Indian Penal Code (IPC) based on the provided input and output format. Please go ahead and provide me with the examples and the necessary information for the case you'd like me to analyze.

    \vspace{1mm}
    \noindent \textbf{INPUT\_PROMPT: }

    \noindent \textbf{Output Format:} 

    List of relevant Sections and their titles

    \noindent \textbf{Examples:}
    
    \textbf{Facts 1: <In-context Facts 1 go here>}

    \textbf{Output 1: <Gold-standard Labels for Facts 1 go here>}

    \textbf{\ldots}

    \textbf{Facts n+1: <Test Facts go here>}
    
    \textbf{Output:}  
\end{alltt} \\\hline
    \end{tabular}
    \caption[LSI Task GPT-3.5 Prompt]{Prompt template for LSI for GPT-3.5 ($n$ in-context examples)}
    \label{prompt:lsi-3.5}
\end{table}

\begin{table}[htbp]
    \tiny
    \centering
    \begin{tabular}{|p{0.9\linewidth}|}\hline
        
       \begin{alltt}

    \vspace{1mm}
    \noindent \textbf{SYSTEM\_PROMPT:} You are an intelligent Legal Statute Identification (LSI) system. You will be provided the the facts of an Indian court document. You need to output the Sections of the Indian Penal Code (IPC) and their corresponding titles which are possibly violated given the facts of the document. You should strictly adhere to the output format. Do not output anything else. It has been found out that on average, each Indian Court Document has may contain between 1 and 12 relevant statutes per document. Keep this in mind while finding out relevant statutes.
    
          \vspace{1mm}
    \noindent \textbf{USER\_PROMPT:} Are you clear about your role?

    \vspace{1mm}
    \noindent \textbf{ASSISTANT\_PROMPT: }Yes, I understand my role as an intelligent Legal Statute Identification (LSI) system. My task is to identify and output the IPC sections and their corresponding titles that may be violated provided the facts of the document. I will adhere strictly to the specified output format. Please provide me with the facts, and I will do my best to assist you.

    \vspace{1mm}
    \noindent \textbf{INPUT\_PROMPT: }

    \noindent \textbf{Output Format:} 

    <Statute 1>

    <Statute 2>

    \ldots

    <Statute x>

    \noindent \textbf{Examples:}
    
    \textbf{Facts 1: <In-context Facts 1 go here>}

    \textbf{Output 1: <Gold-standard Labels for Facts 1 go here>}

    \textbf{\ldots}

    \textbf{Facts n+1: <Test Facts go here>}
    
    \textbf{Output:}  
\end{alltt} \\\hline
    \end{tabular}
    \caption[LSI Task GPT-4 Prompt]{Prompt template for LSI for GPT-4 ($n$ in-context examples)}
    \label{prompt:lsi-4}
\end{table}

\noindent\textbf{Data Selection:}
We randomly chose 100 documents (in this case, fact portions) from the ILSI test set, all of which satisfied the length constraints of GPT.
For ICL, we sample other documents from the test set while satisfying the length constraints.
Also, for IC examples, we collate the gold-standard Section numbers and their respective titles in the form \textit{Section x --- Title of Section x}, create a numbered list, and pass it to GPT.

\begin{table}
    \centering
    \small
    \begin{tabular}{cccc}\toprule
         \textbf{Model}&  \textbf{mP} &  \textbf{mR}& \textbf{mF1}\\ \midrule
         GPT-3.5 0-shot&  21.60&  \textbf{32.55}& 21.55\\
         GPT-3.5 1-shot&  \textbf{27.06}&  22.07& \textbf{22.61}\\
         GPT-3.5 2-shot&  25.35&  21.53& 21.40\\ \midrule
         GPT-4 0-shot&  25.31&  26.74& 23.99\\
         GPT-4 1-shot&  27.13&  23.22& 22.26\\
         GPT-4 2-shot&  25.16&  20.89& 20.53\\ \bottomrule
    \end{tabular}
    \caption[LSI Task GPT Results]{Macro-averaged scores for ILSI dataset}
    \label{tab:lsi_cgpt_results}
\end{table}

\noindent\textbf{Verbalizer:} Due to the flexibility of the output format for GPT-3.5, it can output a lot of Sections from the IPC and even other acts. We filtered the outputs by considering if either the Section number OR the Section title matched with any of the 100 IPC Section numbers and the corresponding titles of the ILSI candidate statute set. The OR condition was necessary since we observed that even with the second setting, GPT still suffers from the hallucination problem, sometimes providing the correct Section titles with non-existent Section numbers. For instance, consider the GPT output \textit{``Section 1565 of the Indian Penal Code (IPC) - Liability of abettor when one act abetted and different act done''}. This is a hallucinated output since IPC does not have more than 600 Sections. But, the title actually corresponds to a Section in IPC, namely Section 111. Although the output format for GPT-4 was stricter, we still had to perform these processing steps so that we could match with either the Section number or the title.

\noindent\textbf{Results:} The ILSI dataset is quite challenging, as seen in the SOTA results. The results for the GPT models are listed in Table~\ref{tab:lsi_cgpt_results}. In such a comparison, GPT-3.5 does not perform too badly, as compared to other tasks. GPT-4 improves upon this score. ICL does not seem to help too much, with 0,1 and 2-shot settings showing very little difference in results for GPT-3.5. However, the performance of GPT-4 actually decreases significantly with ICL, even performing worse than some GPT-3.5 settings.


\subsubsection{Summarization (SUMM)}
\label{app:summ-chatgpt}

\begin{table}[htbp]
    \tiny
    \centering
    \begin{tabular}{|p{0.9\linewidth}|}\hline
        
       \begin{alltt}

    \vspace{1mm}
    \noindent \textbf{SYSTEM\_PROMPT:} You are a smart and intelligent summarization system, trained to read, understand an summarize Indian court case documents. Your task is, given a court case document and a target summary length, generate a detailed summary of the case in your own words within the specified length. The summary should contain ALL the important legal aspects of the case. I will provide you with the document to be summarized.
    
          \vspace{1mm}
    \noindent \textbf{USER\_PROMPT:} Are you clear about your role?

    \vspace{1mm}
    \noindent \textbf{ASSISTANT\_PROMPT:} Sure, I'm ready to help you with your court judgment summarization task. Please provide me with the examples and the case document I'm supposed to summarize.

    \vspace{1mm}
    \noindent \textbf{INPUT\_PROMPT: }

    \noindent \textbf{Output Format:} Generate the summary in a few simple paragraphs. Do not use any paragraph headers, bullet points, or any other such formatting.

    \noindent \textbf{Examples:}
    
     Case Document 1: \textbf{<In-context Document 1 goes here>}

    Summary 1 (in \textbf{<Length of reference summary goes here>} words): \textbf{<Reference summary for Document 1 goes here>}

    \textbf{\ldots}

    Case Document n+1: \textbf{<Test Document goes here>}
    
    Summary n+1 (in \textbf{<Length of reference summary goes here>} words): 
\end{alltt} \\\hline
    \end{tabular}
    \caption[SUMM Task GPT Prompt]{Prompt template for SUMM for both GPT-3.5 and GPT-4 ($n$ in-context examples)}
    \label{prompt:summ}
\end{table}


\noindent\textbf{Prompt Design:}
{GPT is known to be more conversant with the abstractive summarization task. Hence, we provide the model with the summary length limit and ask it to generate the summary (see Table~\ref{prompt:summ}). 
A large majority of the judgments (more than 95\%) can be passed as a whole to GPT-3.5. For the rest of the (longer) documents, we break the documents into two chunks, summarize each chunk individually, and then append the chunk summaries to get the final summary. For GPT-4, all documents can be passed without chunking.}

\noindent\textbf{Data Selection:}
{We chose all 100 documents from the test set of In-Abs for passing to ChatGPT. For ICL, we sample from this set of documents itself. We try to sample the smallest samples for the longer input examples to fit the entire prompt within GPT token length limit.}

\noindent\textbf{Verbalizer:} The entire output returned by GPT is considered as the abstractive summary. We specifically instruct GPT to output simple text, without headers and bullet points. We observe that both GPT-3.5 and GPT-4 mostly adhere to this instruction.

\begin{table}
    \centering
    \small
    \setlength\tabcolsep{5pt}
    \begin{tabular}{ccccc}\toprule
         \textbf{Model}&  \textbf{R-1} &  \textbf{R-2}& \textbf{R-L} &\textbf{BERTScore}\\ \midrule
         GPT-3.5 0-shot &  0.392&  0.165& 0.208&0.847\\
         GPT-3.5 1-shot&  0.385&  0.141& 0.201&0.835\\
         GPT-3.5 2-shot&  0.419&  0.164& 0.215&0.838\\ \midrule
        GPT-4 0-shot&  \textbf{0.472}&  \textbf{0.183}& \textbf{0.228}&\textbf{0.848}\\
         GPT-4 1-shot&  0.304&  0.059& 0.161&0.807\\
         GPT-4 2-shot&  0.324&  0.080& 0.171&0.813\\\bottomrule
    \end{tabular}
    \caption[SUMM Task GPT Results]{Rouge-1,2,L and BERTScore scores for SUMM}
    \label{tab:summ_cgpt_results}
\end{table}

\noindent\textbf{Results:}
GPT results are shown in Table~\ref{tab:summ-results}. Even for this task (which GPT is quite conversant with), the gap in performance compared to the SOTA is significant. GPT-4 performs slightly better than GPT-3.5 in a zero-shot setting. For both models, 1-shot prompting reduces the performance, although the difference is much larger for GPT-4, which performs very poorly in a 1-shot setting. On using 2-shot setting, we observe an improvement again, but it is not too much (for GPT-4, 2-shot still performs significantly poorly compared to 0-shot). It is possible that the IC examples are actually confusing the model rather than helping it.

\subsubsection{Legal Machine Translation (L-MT)}
\label{app:lmt-chatgpt}

\noindent\textbf{Prompt Design:}
{GPT is known to perform translations effectively. Hence, we provide the model with just the input sentence (in English), and we ask the model to translate the sentence to the desired target language (see Table~\ref{prompt:lmt}).}

\noindent\textbf{Data Selection:}
{We randomly choose 5 samples from each target language from each MILPaC dataset. This gives us 45 documents each for MILPaC-IP and MILPaC-Acts (9 target languages), and 20 documents for MILPaC-CCI-FAQ (4 target languages), giving us a total of 110 samples. It should be noted that all datasets contain two types of samples -- questions and answers. However, the answers from the MILPaC-CCI-FAQ dataset consist of just a single number corresponding to different choices in the MCQ setting. Thus, we do not choose the answer samples from MILPaC-CCI-FAQ.
For ICL, we randomly choose samples from the same target language in the same dataset.}

\begin{table}[htbp]
    \tiny
    \centering
    \begin{tabular}{|p{0.9\linewidth}|}\hline
        
       \begin{alltt}

    \vspace{1mm}
    \noindent \textbf{SYSTEM\_PROMPT:} You are a smart and intelligent machine translation system, trained to read Indian legal texts and translate them to Indian languages. Your task is, given an English language sentence from a legal document, translate it to the given target Indian language. I will provide you with the input/output format, target language and the sentence to be translated. I will also provide some examples of the task.
    
          \vspace{1mm}
    \noindent \textbf{USER\_PROMPT:} Are you clear about your role?

    \vspace{1mm}
    \noindent \textbf{ASSISTANT\_PROMPT:} Sure, I'm ready to help you with your legal translation task. Please provide me with the sentence and the target language I am supposed to translate to.

    \vspace{1mm}
    \noindent \textbf{INPUT\_PROMPT: }

    \noindent \textbf{Examples:}
    
    \textbf{Sentence 1 in English: <In-context Sentence 1 goes here>}

    \textbf{Sentence 1 in <Target language goes here>: <Reference translation for Sentence 1 goes here>}

    \textbf{\ldots}

    \textbf{Sentence n+1 in English: <Test Document goes here>}
    
    \textbf{Sentence n+1 in <Target language goes here>:}  
\end{alltt} \\\hline
    \end{tabular}
    \caption[L-MT Task GPT Prompt]{Prompt template for L-MT for both GPT-3.5 and GPT-4(for $n$ in-context examples)}
    \label{prompt:lmt}
\end{table}

\noindent\textbf{Verbalization:} We directly take the entire GPT output as the translation.

\begin{table*}
\small
\setlength\tabcolsep{3pt}
    \centering
    \begin{tabular}{cccccccc} \toprule
         \multirow{2}{*}{Dataset}&  \multirow{2}{*}{\# Shots} &  \multicolumn{3}{c}{GPT-3.5} &  \multicolumn{3}{c}{GPT-4}\\ 
         & & BLEU & GLEU & chrF++ & BLEU & GLEU & chrF++ \\ \midrule
         MILPac-IP&  0&  26.2&  30.3& 45.3 & 36.3 & 39.2 & 53.5\\
         &  1&  27.8&  31.5& 46.3 & 37.6 & 40.5 & 54.0 \\
         &  2&  27.9&  31.0& 45.4 & \textbf{38.0} & \textbf{40.6} & \textbf{54.5}\\\midrule
         MILPaC-CCI-FAQ&  0&  24.1&  28.2& 43.9 & 35.0 & 36.1 & 50.0\\
         &  1&  25.9&  28.7& 43.8 & 39.0 & 41.4 & 56.6 \\
         &  2&  27.9&  30.6& 44.9 & \textbf{42.2} & \textbf{43.3} & \textbf{57.2}\\\midrule
         MILPaC-Acts&  0&  18.2&  23.1& 36.0 & 29.0 & \textbf{32.6} & 45.7 \\
         &  1&  19.5&  23.6& 36.6 & 28.8 & 32.3 & 45.3\\
         &  2&  21.2&  24.8& 38.2 & \textbf{29.1} & 32.4 & \textbf{46.2} \\\midrule
 Average& 0& 22.8& 28.2&43.9 & 33.4 & 36.1 & 50.0 \\
 & 1& 24.4& 27.9&42.3 & 35.1 & 38.0 & 52.0 \\
 & 2& 25.6& 28.8&42.8 & \textbf{36.4} & \textbf{38.7} & \textbf{52.6}\\\bottomrule
    \end{tabular}
    \caption[L-MT Task GPT Results]{Corpus-level BLEU, GLEU, and chrF++ scores for GPT-3.5 and GPT-4 prompting with 0, 1 and 2 shot settings}
    \label{tab:lmt_cgpt_results}
\end{table*}


\noindent\textbf{Results:} GPT-3.5 produces decent results for L-MT as compared to SOTA approaches, possibly due to GPT's prior knowledge on this task. This is further improved upon by GPT-4, which actually outperforms SOTA on average for two of three metrics. For both models, there is a drop in performance for Acts, possibly due to the more complex nature of the text in the Acts dataset~\citep{mahapatra2023milpac}. We see a gradual improvement across all metrics and all datasets for both GPT-3.5 and GPT-4 with an increasing degree of ICL, with 2-shot prompting producing the best results.

\subsection{Experiments with Smaller LLMs}

In addition, we also experimented with other large language models with smaller parameter sizes. Specifically, we experimented with GPT-Neo \cite{gpt-neo} family of three models (GPT-Neo-125M, GPT-Neo-1.3B, GPT-Neo-2.7B) trained on the Pile dataset \cite{pile}, GPT-J-6B \cite{gpt-j}, Llama-2-7b-chat-hf \cite{touvron2023llama}, and recently released Mistral-7B-v0.1 \cite{jiang2023mistral} language models for our experiments. The primary challenge when validating the smaller language model is the prompt design. Following previous works \cite{brown2020language-incontext-learning,robinson2023leveraging}, we pose the prompt in a multiple-choice question-answering format (a prompt sample for various tasks present in the benchmark can be found in the supplementary material) and validate the performance using the obtained log probability of the predicted tokens as highlighted in \cite{robinson2023leveraging}. Moreover, since the tasks are more complicated with larger context lengths, the generative models sometimes generate some irrelevant tokens. For those cases with random token generation, we consider it to be a failure case and use a random prediction as a proxy of predictions. Overall, we observed that all the language models perform poorly with near-random predictions over the proposed set of legal language understanding tasks. 

We speculate two primary reasons for this finding. First, the language models we used are not explicitly designed to capture the question-answering format for a larger context. Since the context length of the task in the proposed benchmark is significantly higher than the other NLU tasks, it becomes more challenging for smaller language models to decode the question-answer format required for performing these tasks.
Second, these models lack the instruction tuning strategies followed by larger models like GPT3.5, making it much harder to capture the context. Moreover, our experiments with GPT3.5 also suggest that if the context is large, even the larger models fail to capture the requested instructions present in the query prompt.

\end{document}